\documentclass[10pt,twocolumn,letterpaper]{article}

\usepackage{wacv}
\usepackage{times}
\usepackage{epsfig}
\usepackage{graphicx}
\usepackage{amsmath}
\usepackage{amssymb}

\usepackage{color}
\usepackage{comment}
\usepackage{multirow}
\usepackage{multicol}
\usepackage{subfig}
\usepackage{booktabs}
\usepackage{caption}

\def\wacvPaperID{754} % Enter the WACV Paper ID here

\wacvfinalcopy % *** Uncomment this line for the final submission
\ifwacvfinal
\def\assignedStartPage{9876} % *** Enter the assigned starting page number (instead of 9876)
\fi

\ifwacvfinal
\usepackage[breaklinks=true,bookmarks=false]{hyperref}
\else
\usepackage[pagebackref=true,breaklinks=true,colorlinks,bookmarks=false]{hyperref}
\fi

\ifwacvfinal
\else
\fi

\usepackage[firstpage]{draftwatermark}
\SetWatermarkFontSize{1cm}
\SetWatermarkAngle{90}
\usepackage[printwatermark]{xwatermark}
\usepackage{xcolor}

\begin{document}
% watermark
%\newwatermark[pagex={1},fontfamily=bch,fontsize=11pt,color=gray!60,angle=60,scale=1,xpos=0,ypos=14cm]{Published in WACV 2021}
\SetWatermarkText{WACV 2021}

%%%%%%%%% TITLE
\title{Coarse Temporal Attention Network (CTA-Net) for Driver's Activity Recognition}

\author{Zachary Wharton \quad Ardhendu Behera \quad Yonghuai Liu \quad Nik Bessis\\
Edge Hill University, Ormskirk, Lancashire, United Kingdom\\
{\tt\small zachary.wharton@go.edgehill.ac.uk, \{beheraa, liuyo, bessisn\}@edgehill.ac.uk}
% For a paper whose authors are all at the same institution,
% omit the following lines up until the closing ``}''.
% Additional authors and addresses can be added with ``\and'',
% just like the second author.
% To save space, use either the email address or home page, not both
% \and
% Second Author\\
% Institution2\\
% First line of institution2 address\\
% {\tt\small secondauthor@i2.org}
}

\maketitle
%\thispagestyle{empty}

%%%%%%%%% ABSTRACT
\begin{abstract}
 There is significant progress in recognizing traditional human activities from videos focusing on highly distinctive actions involving discriminative body movements, body-object and/or human-human interactions. Driver's activities are different since they are executed by the same subject with similar body parts movements, resulting in subtle changes. To address this, we propose a novel framework by exploiting the spatiotemporal attention to model the subtle changes. Our model is named Coarse Temporal Attention Network (CTA-Net), in which coarse temporal branches are introduced in a trainable glimpse network. The goal is to allow the glimpse to capture high-level temporal relationships, such as ‘during’, ‘before’ and ‘after’ by focusing on a specific part of a video. These branches also respect the topology of the temporal dynamics in the video, ensuring that different branches learn meaningful spatial and temporal changes. The model then uses an innovative attention mechanism to generate high-level action specific contextual information for activity recognition by exploring the hidden states of an LSTM. The attention mechanism helps in learning to decide the importance of each hidden state for the recognition task by weighing them when constructing the representation of the video. Our approach is evaluated on four publicly accessible datasets and significantly outperforms the state-of-the-art by a considerable margin with only RGB video as input.
\end{abstract}
%%%%%%%%% BODY TEXT
\vspace{-.5cm}
\section{Introduction}
Recognizing human/driver activities while driving is not only a key ingredient for the development of Advanced Driver Assistance System (ADAS) but also for the development of many intelligent transportation systems. These include autonomous driving \cite{merat2014transition,Kim17}, driving safety monitoring \cite{prat17,Kaplan15}, Vehicle to Vehicle (V2V) and Vehicle to Infrastructure (V2I) \cite{talebpour2016modeling} systems, just to name a few. The rise of automation and a growing interest in fully autonomous vehicles encourage more non-driving or distractive behaviors of the driver. Therefore, understanding human drivers' behavior is crucial for accurate prediction of Take-Over-Request and surrounding vehicles' activities, which result in developing control strategies and human-like planning. Moreover, understanding drivers' behavior such as human drivers' interaction with each other, as well as with transportation infrastructure provides significant insight into the efficient design of V2V and V2I systems. Similarly, real-time monitoring of drivers' activities and body language constitutes a safe driving profile for each driver. It is vital for emerging vehicle/ride sharing industries and fleet management platforms.

%It is expected that automation will reduce the driver's role. However, the vehicle also requires to deliver human-driver like performance if it is to be trusted \cite{merat2014transition,Kim17}. %This could lead to over-reliance on automation and might lead to catastrophic consequences. Therefore, the driver is required to be intervened in case of uncertainties \cite{ohn2016looking}. 
Real-world driving scenarios are a multi-agent system in which diverse participants interact with each other and with infrastructures. Moreover, each driver has their own driving style and often depends on sophisticated multi-tasking human intelligence, including the perception of traffic situations, reasoning surrounding road-users' intentions, paying attention to the  potential hazards, planning ego-trajectory, and finally executing the driving task. Therefore, it is a complex problem involving a large diversity in daily driving scenarios, driving behaviors, and different granularity of activities, resulting in significant challenges in understanding and representing driving behaviors. %Toward a complete understanding of driver's behavior, 
To address this, recent research on recognizing fundamental fine-grained driver's actions such as eating, drinking, interacting with the vehicle controls, and so on is only the first step  \cite{martin2019drive,behera2018context,abouelnaga2017real,eraqi2019driver}. 
%There has been also a some progress on recognizing drivers' distraction and drowsiness by analyzing head pose, facial expression, state of the eyes and faces for monitoring of safe driving \cite{Kaplan15}\cite{ohn2016looking}\cite{gao2014detecting}\cite{li2013analysis}. However, little progress has been made in recognizing human driver activity, which often involves human-object and human-car interactions. This could be due to the lack of publicly available datasets for developing model to recognize such interactions. Lately, effort has been made in developing few datasets \cite{martin2019drive}\cite{abouelnaga2017real}\cite{eraqi2019driver} and encourage researchers to develop driver activity monitoring models.   
%On the other hand, recognizing %As a first step, Most of the vision-based research focus on understanding outside environment of the vehicle \cite{cordts2016cityscapes}\cite{yu2018bdd100k}\cite{ramanishka2018toward}\cite{geiger2013vision}\cite{maddern20171} involves detecting and tracking of objects (e.g. vehicles, people, bikes, etc), road sign/landmark detection, road layout, etc. 

%In this work, we focus on recognizing the driver's behavior by analyzing video data. 
Driver behavior recognition is closely linked to the broader field of human action recognition, which has rapidly gained much attention due to the rise of deep learning \cite{hussein2019timeception,hussein2019videograph,piergiovanni2019evolving,wang2018non,carreira2017quo,girdhar2017attentional,tran2018closer}. These approaches are data-intensive and are trained on large-scale video datasets, usually originated from YouTube \cite{carreira2017quo,karpathy2014large}, and consist of highly discriminative actions often executed by different subjects. Whereas, driving behavior commonly involves various driving/non-driving activities executed by the same driver with very similar body parts movements, resulting in subtle changes. % among various activities. 
For example, talking vs texting using a mobile phone, eating vs drinking, etc. in which many actions have a similar upper-body pose and the only difference is the object of interest. Furthermore, in such scenarios, only the part of the body (e.g. upper-body) is visible, making the problem even harder. % for the recognition algorithm. 
Therefore, the above-mentioned conventional human action recognition models might not be suitable for drivers' activities.
%To address this problem, recently visual attention models \cite{baradel2018human}\cite{girdhar2017attentional}\cite{mnih2014recurrent}\cite{sharma2015action} have drawn considerable interest. Such models are able to focus visual cues specific to a given task. 

\noindent\textbf{Our work:} Our CTA-Net uses visual attention in an innovative way to capture both subtle spatiotemporal changes and coarse temporal relationships. It attends \textit{visual cues specific to  temporal segments} to preserve the temporal ordering in a given video and then a temporal attention mechanism, which dictates how much to \textit{attend} the \textit{current visual cues conditioned on their temporal neighborhood contexts}. It is a recurrent model (an LSTM) in which a visual representation of a video frame is learned using a residual network \cite{He16} (ResNet-50). The last convolutional block (\texttt{CONV5}) of our model focuses on a segment of the input video, allowing our novel attention to assign estimated importance to each segment of the video by considering the knowledge of the coarse temporal range. For example, such coarse temporal range might indicate that the driver's hand moving towards an object of interest (e.g. phone, bottle, etc.), carrying out the required task (e.g. talking, drinking, etc.) and then the hand moving away. Many different activities exhibit the same spatiotemporal pattern of the hand moving toward and moving away. However, the proposed coarse temporal range, their temporal ordering, and the appearance of a specific object(s) in a given activity would allow to discriminate different activities. %Moreover, the model also explores the fine-grained (sequence of frames) temporal modeling after the coarse temporal segment using an LSTM to capture the sequential dependencies. 
Moreover, our novel temporal attention \textit{learns to attend} the different parts of the hidden states of the LSTM in discriminating fine-grained activities. 

\noindent\textbf{Our contributions:} They can be summarized as: 1) a driver activity recognition model is proposed with a residual CNN-based glimpse sensor and a novel attention mechanism; 2) our novel attention mechanism is designed to learn how to emphasize the hidden states of an LSTM in an adaptive way; 3) to capture task-specific high-level features, a spatial attention mechanism conditioned on coarse temporal segments is developed by introducing branches in the last convolutional layer; and 4) extensive validation of the proposed model on four datasets, obtaining state-of-the-art results.
%There has been also a recent progress on recognizing drivers' distraction and drowsiness by analyzing head pose, facial expression, state of the eyes and faces for monitoring of safe driving. However, little progress has been made in recognizing human driver activity, which often involves human-object and human-car interactions. This could be due to the lack of publicly available datasets for developing model for recognizing such interactions. Lately, this has been addressed by the proper  
%
%The rest of the paper is organized as follows. In section \ref{sec:rel}, we discuss the related work. The proposed CTA-Net is described in section \ref{sec:proposed}. The experimental evaluations is presented in section \ref{sec:exp}. Finally, we conclude in section \ref{sec:con}.
%
%-------------------------------------------------------------------------
%
%\vspace{-.3cm}
\begin{figure*}[t]
%\begin{center}
\hspace{.6cm}
\begin{minipage}[b]{.39\textwidth}
  \subfloat
    [Glimpse Sensor]
    {\label{fig:figA}\includegraphics[width=\textwidth]{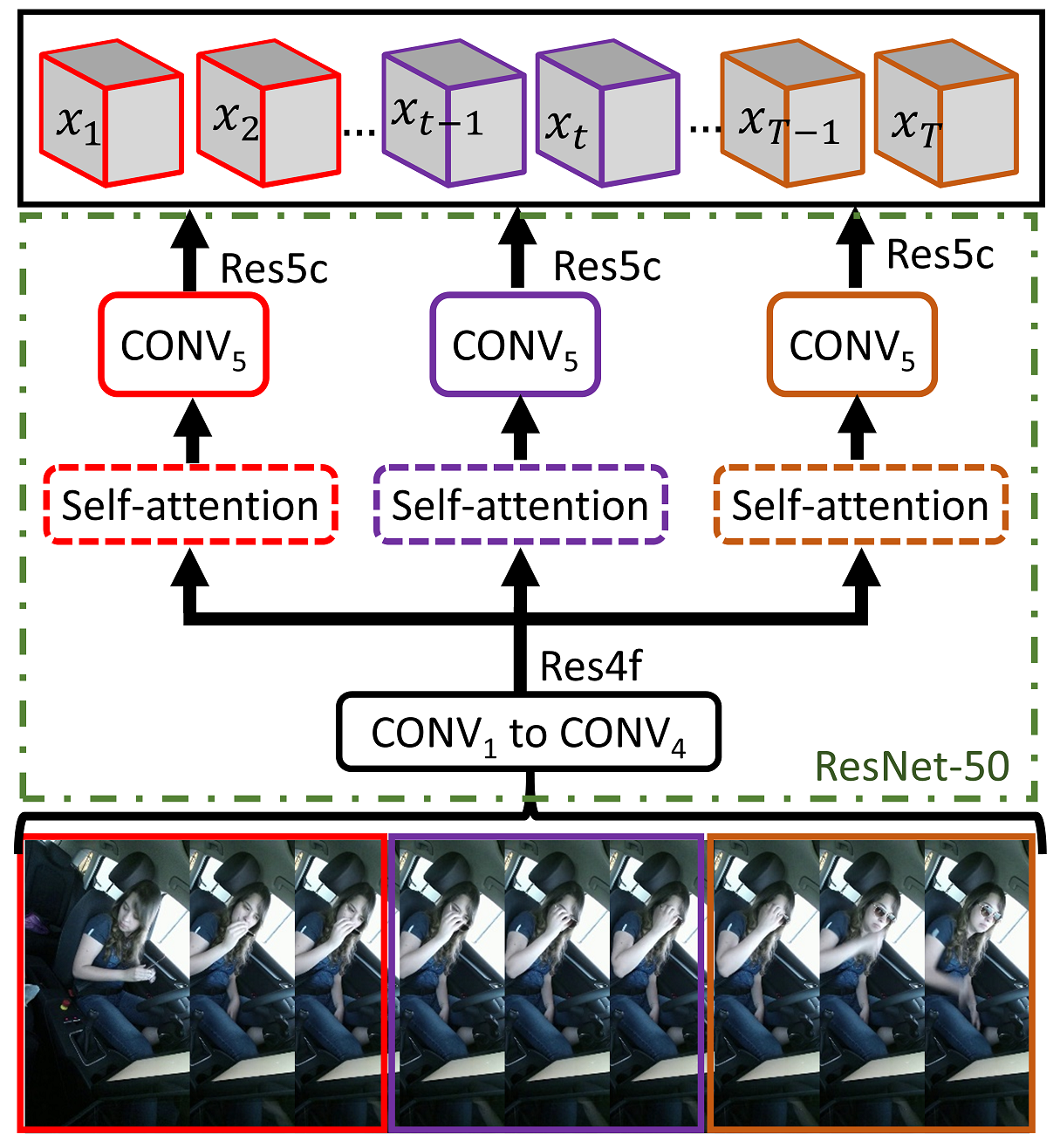}}
  \end{minipage}
\begin{minipage}[b]{.465\textwidth}
\centering
\subfloat
  [Self-Attention layer in glimpse sensor]
  {\label{fig:figB}\includegraphics[width=\textwidth]{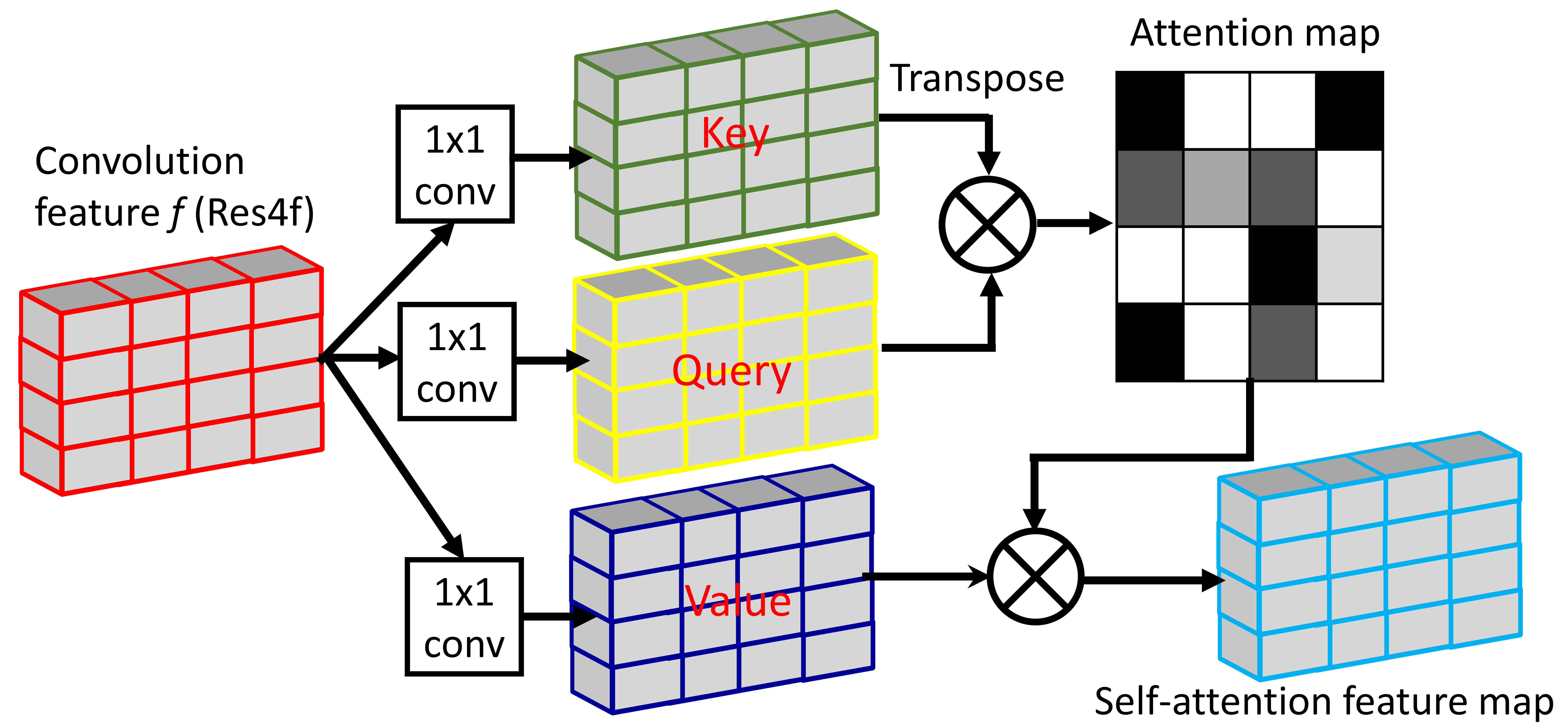}}  
\vfill
\subfloat
  [LSTM and our novel temporal attention]
  {\label{fig:figC}\includegraphics[width=\textwidth]{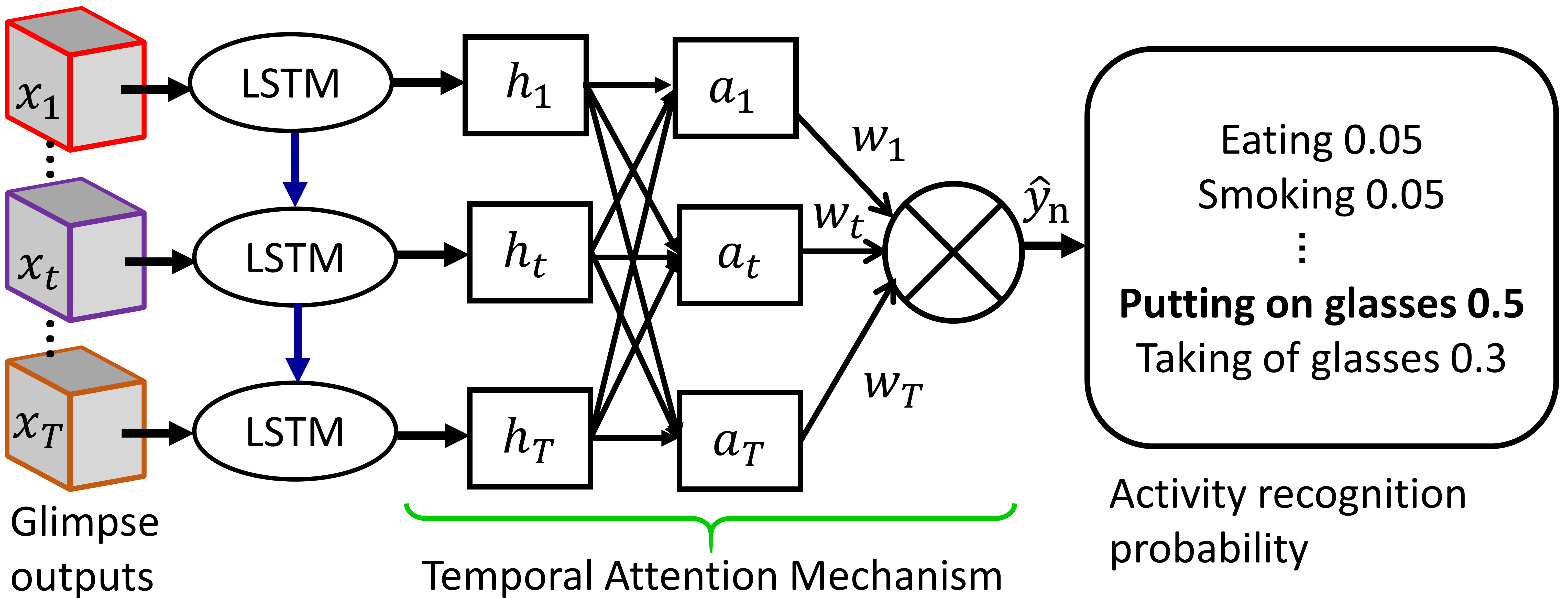}}
\end{minipage}
\caption{The proposed CTA-Net consists of - \textbf{a) Glimpse sensor:} Given an input video $v$ consisting $T$ frames, the sensor $f_g(.;\theta_g)$ extracts feature $x_t$ of the $t^{th}$ frame, where $t=1\dots T$. \textbf{b) Self-Attention:} It captures important cues on activity-specific spatial changes. % during the extraction of \textit{retina-like} representation by considering large receptive fields. %The layer is adapted from SAGAN \cite{zhang2018self} in which the \textit{query}, the \textit{key} and the \textit{value} are all the same. 
\textbf{c) Temporal Attention:} The module $f_a(.;\theta_a)$ uses the internal state $h_t$ of an LSTM $f_h(.;\theta_h)$ (unrolled) that takes as input $x_t$ and selectively focuses on the $h_t$ to infer activity.}
\label{fig:main}
%\end{center}
\vspace{-.4cm}
\end{figure*}
\section{Related Work and Motivation}\label{sec:rel}
\noindent \textbf{Traditional Human Activity Recognition:} Recent surge of deep learning has significantly influenced the advancement in recognizing human activities from videos. % there has been active research dedicated to the development of deep networks for videos. 
Most attempts in this genre are usually derived from the image-based networks, which are used to extract features from individual frames and extended them to perform temporal integration by forming a fixed size descriptor using statistical pooling such as max and average pooling \cite{hussein2017unified,habibian2016video2vec}, attentional pooling \cite{girdhar2017attentional}, rank pooling \cite{fernando2016rank}, context gating \cite{miech2017learnable} and high-dimensional feature encoding \cite{girdhar2017actionvlad,xu2015discriminative}. However, an important visual cue representing the temporal pattern is overlooked in such statistical pooling and high-dimensional encoding. 
On the other hand, recurrent networks \cite{behera2018context,Xu_2019_ICCV}, Temporal Convolutional Networks (TCN) \cite{lea2017temporal}, and learning spatiotemporal features through 3D convolutions \cite{tran2018closer,piergiovanni2019evolving,carreira2017quo} are used to capture temporal dependencies. Recurrent networks such as LSTMs are capable of modeling long-term dependencies and thus, adapted in the activity recognition problem. %, which is temporally recurrent by nature. 
To the best of our knowledge, no substantial improvements have been reported recently.

To learn long-term temporal dependencies, Hussein et al. propose Timeception \cite{hussein2019timeception}, which uses multi-scale temporal convolutions to reduce the complexity of 3D convolutions. In \cite{wang2018non}, Wang et al.  present non-local operations as a generic family of building blocks for capturing long-range dependencies. %Sigurdsson et al. \cite{sigurdsson2017asynchronous} use CRF on top of the CNN feature maps. 
Zhou et al. \cite{zhou2018temporal} introduce a Temporal Relation Network (TRN) to learn and reason about temporal dependencies between video frames at multiple time scales. 
Similarly, Wang et al. \cite{wang2016temporal} propose a Temporal Segment Network (TSN) with a sparse temporal sampling strategy. % and video-level supervision to enable learning using the whole action video. 
A Long-term Temporal Convolution (LTC) is proposed  in \cite{varol2017long} to consider different temporal resolutions as a substitute to bigger temporal windows.
%More recently, a multi-scale temporal convolutions called Timeception is proposed by Hussein et al. \cite{hussein2019timeception} to reason about minute-long temporal pattern. 
Another influential approach is the use of 3D CNNs for action recognition. Carreira and Zisserman \cite{carreira2017quo} propose a model (I3D) that %holds the best action recognition results on the large-scale Kinetics dataset. The model 
inflates 2D CNNs pre-trained on images to 3D for video classification. Tran et al. \cite{tran2018closer} describe a spatiotemporal convolution by factorizing the 3D convolutional filters into separate spatial and temporal components to recognize actions. %Similarly, Piergiovanni et al. \cite{piergiovanni2019evolving} propose an evolving space-time neural architecture, which allows learning of space-time interactions.% over a longer time for efficient video recognition.

\noindent \textbf{Attention in Activity Recognition:} Attention mechanism in machine learning has drawn increasing interest in areas such as %machine translation \cite{vaswani2017attention}, 
video question answering \cite{li2019beyond}, video captioning \cite{pei2019memory,song2017hierarchical}, and video recognition \cite{girdhar2017attentional,girdhar2019video,baradel2018human,song2017end,sharma2015action}. This is influenced by human perception, which focuses selectively on parts of the scene to acquire information at specific places and times. %Therefore, integrating attention has the potential to improve the overall accuracy since the model could focus selectively on parts of the data. 
This has been explored by Girdhar and Ramanan \cite{girdhar2017attentional} for action recognition by bottom-up and top-down attention. % as a low-rank approximation of bilinear pooling. 
Similarly, a recurrent mechanism is proposed in \cite{sharma2015action}, focusing selectively on the part of the video frames, both spatially and temporally. % for action recognition. 
Girdhar et al. \cite{girdhar2019video} propose an attention mechanism that learns to emphasize hands and faces to discriminate an action. 
An LSTM-based temporal attention mechanism is proposed by Baradel et al. \cite{baradel2018human} to emphasize features representing hands. % in which the model learns features representing four hands over time to recognize human activities.  
Song et al. \cite{song2017hierarchical} propose an end-to-end spatial and temporal attention to selectively focus on discriminative skeleton joints in each frame and pays different levels of attention to the frames.

\noindent \textbf{Driver Activity Recognition:} Driver activities are a subset of conventional human activities \cite{martin2019drive,behera2018context,abouelnaga2017real,eraqi2019driver,martin2018body,oliver2000graphical,doshi2011tactical,ramanishka2018toward}. It can be categorized into two sub-classes: 1) primary maneuvering (e.g. passing, changing lanes, start, stop, etc.) \cite{oliver2000graphical,doshi2011tactical,ramanishka2018toward} and 2) secondary non-driving (e.g. eating, drinking, talking, etc.) \cite{martin2019drive,behera2018context,abouelnaga2017real,eraqi2019driver,martin2018body} activities. In this work, we focus on secondary activities, which are crucial for safe driving and take-over-request. Moreover, it will be more frequent during the autonomous driving mode. Martin et al. \cite{martin2018body} propose a method to combine multiple streams involving body pose and contextual information. % to recognize driver's secondary activities. 
Behera et al. \cite{behera2018context} advocate a multi-stream LSTM for recognizing driver's activities by combining high-level body pose and body-object interaction with CNN features. A genetically weighted ensemble approach %to combine five different CNNs 
is used in %by Abouelnaga et al. 
\cite{abouelnaga2017real}. % to recognize driver's state. 
The VGG-16 \cite{simonyan2014very} network is modified by Baheti et al. \cite{Baheti18} to reduce the number of parameters for faster execution. Similarly, Li et al. \cite{li2019learning} propose a tactical behavior model that explores the egocentric spatial-temporal interactions to understand how human drives and interacts with road users.   
\newline
\noindent \textbf{Motivation:} It is evident that the traditional activity recognition models are developed to recognize highly distinctive actions. % involving discriminative body movements, body-object and/or human-human interactions. 
Lately, attention mechanisms are brought in to improve the recognition accuracy of these models. The conventional models are adapted for drivers' activity monitoring by tweaking a few layers %without changing network structure 
or simply evaluating the target driving datasets. %Furthermore, a common observation is that many of these models are focused on either recognizing long-term temporal dependencies \cite{sigurdsson2017asynchronous,zhou2018temporal,wang2016temporal} or spatiotemporal convolutions \cite{carreira2017quo,tran2018closer} to capture fine-grained changes. To the best of our knowledge, no existing approaches combine them to recognize human activities. 
In this work, we move a step forward by innovating within frame self-attention, between frames coarse and fine-grained temporal attention to recognize driver's secondary activities. These activities are different from the traditional human activities since they are executed by the same subject resulting in subtle changes among various activities. %Our work is inspired by the TSN \cite{wang2016temporal} and the action transformer network \cite{girdhar2019video}. %This paper addresses this by %revisiting many of the aforementioned approaches (specifically attention, recurrent models and LSTM) in the context of an empirical analysis focuses on understanding the effects of 
%. First, even though our glimpse sensor is similar to visual attention in \cite{mnih2014recurrent}, %which uses different image patch size from a given location to model visual attention in an image. Whereas, 
Our coarse temporal attention introduces three branches to model high-level temporal relationships (`during', `before', and `after') with the assumption is that main action is performed in `during' (e.g. drinking), `before' focuses on pre-action event (e.g. take the bottle) and `after' emphasizes on post-action episode (e.g. put the bottle). % instead of image patch size from a given location to model attention in images \cite{mnih2014recurrent}. 
The self-attention within each branch selectively focuses on capturing spatial changes. % involving long-range and multi-level dependencies. 
Finally, we introduce a novel temporal attention by focusing on the distribution of hidden states of an LSTM instead of image feature maps \cite{girdhar2017attentional} or hard attention involving the subject's hands \cite{baradel2018human}. We argue that our contribution includes not only the design of the CTA-Net but also an empirical study on the role of attention in improving accuracy.
\vspace{-.2cm}
\section{Proposed End-to-End CTA-Net}\label{sec:proposed}
\subsection{Problem formulation} \label{sec:prob}
For video-based activity recognition, we are given $N$ training videos $V = \{v_n|n=1\dots N\}$ and the activity label $y_n$ for each video $v_n$. %, where $N$ is the total number of videos. 
The aim is to find a function $F$ that predicts $\hat{y}=F(v)$ that matches the actual activity $y$ of a given video $v$ as much as possible. We learn $F$ by minimizing the categorical cross-entropy $L_v$ between the predicted $\hat{y}_n$ and the actual activity $y_n$:
\begin{equation}
    L_v = -\sum_{n=1}^{N}y_n \text{log}(\hat{y}_n), \text{where } \hat{y}_n = F(v_n)
\end{equation}
%\vspace{-0.1cm}
\subsection{Glimpse sensor} \label{sec:glimpse}
The CTA-Net is built around glimpse sensor for visual attention \cite{mnih2014recurrent} in which information in an image is adaptively selected via %a bandwidth-limited sensor by focusing on locations of interest. The sensor encodes
encoding regions progressively around a given location in the image. Inspired by this, our approach encodes information in temporal locations within a video. The proposed glimpse $f_g$ receives image $I_t$ ($t=1\dots T$) at time $t$ from a video $v_n$. %It extracts a feature vector $x_t$ from $I_t$ by limiting the temporal bandwidth around %the temporal position $t$ of the video $v_n$. The proposed glimpse sensor $f_g$ 
It produces the glimpse feature vector $x_t=f_g(I_t,t_c;\theta_g)$ from $I_t$ by limiting the temporal bandwidth around $t$, where $t_c$ is the coarse temporal bandwidth of the video $v_n$ and $\theta_g$ is the model parameter.

Our glimpse is implemented using ResNet-50 \cite{He16} (Fig. \ref{fig:figA}). % due to its powerful representational ability resulting in superior performance in solving image recognition problems. 
We modify this network by introducing two essential ingredients: 1) Coarse temporal bandwidth $t_c$ and 2) Self-Attention layer (Fig. \ref{fig:figB}). The $t_c$ aims to limit $f_g$ to focus on certain temporal positions in $v_n$. If it is limited to a single frame (i.e. $t_c=1$) then the sensor complexity will increase. % (e.g. a large number of parameters). 
To address this, we use coarse bandwidth ($t_c=T/3$). It allows $f_g$ to focus on different temporal parts of a video, motivated by %the approach in
\cite{behera2014real} that uses \textit{before}, \textit{during}  and \textit{after} to capture the temporal relationships in a video. Moreover, driver secondary activities often involve human-object interactions (e.g. phones, car controls, etc.) and consist of spatiotemporal dynamics such as: i) hand approaching towards objects, ii) object manipulation, and iii) hand moving away. This involves three distinctive sub-activities. Our approach explores it by introducing three branches involving the last $CONV_5$ block of ResNet-50 (Fig. \ref{fig:figA}). The reason is that CNNs learn features from general (e.g. color blobs, Gabor filters, etc.) to more specific (e.g. shape, complex structures, etc.) as we move from the input to output layer. Thus, we share the parameters of lower layers ($CONV_1$ to $CONV_4$) among frames to produce a generic representation that is then processed by the bandwidth-specific layers ($t_c$, where $c=1,2,3$) to generate the required outputs.  

Within each branch of $f_g$ (Fig. \ref{fig:figA}), we also add an attention map $\theta_p$ (Fig. \ref{fig:figB}) to capture bandwidth-specific important cues focusing on spatial changes. % in lower-resolution feature maps. 
The aim is to model long-range, multi-level dependencies across image regions and is complementary to the convolutions to capture the spatial structure within the image. Our model explicitly learns the relationships between features located at $i^{th}$ and $j^{th}$ position in $f_g$ and is represented as $p(f_g^{i}|f_g^{j};\theta_p)$, $\forall i$ $i\ne j$. It conveys how much to focus on the $i^{th}$ location when synthesizing the $j^{th}$ position in $f_g$. To achieve this, we compute the attention map $\theta_p$ by adapting the self-attention in SAGAN \cite{zhang2018self} where the \textit{query}, the \textit{key} and the \textit{value} are computed from feature map $Res4f$ (Fig \ref{fig:figB}) via three separate $1\times 1$ convolutions. % separately using  %block of ResNet-50 . 
The \textit{key} multiplies with the \textit{query} and %undergo a matrix multiplication and %then . The result
then use a softmax to create attention map $\theta_p$. %a probability distribution (i.e. attention map).
The \textit{value} is multiplied with $\theta_p$ to get the desired output $o_c$ ($c=1,2,3$). %Like in \cite{zhang2018self}, we use the spectral normalization \cite{miyato2018spectral} in all $1\times 1$ convolutions and 
Afterwards for each frame at $t$, %The final output is computed by multiplying 
$o_c$ is multiplied with a learnable scalar $\gamma$ (initialized as zero) and added back to the input as a residual connection, i.e. $\hat{o}_{c,t}=o_c*\gamma + Res4f_t$. %, where $Res4f_t$ is the convolution feature map of frame $I_t$. 
The feature map $\hat{o}_{c,t}$ passes through the $CONV_5$ block (Fig. \ref{fig:figA}) to produce the desired glimpse feature vector $x_t$.     
%
%
%\vspace{-0.1cm}
\begin{figure*}[t]
    \centering
    \subfloat[Drive\&Act
    \cite{martin2019drive}]{\includegraphics[width=0.11\textwidth,height=0.12\textwidth]{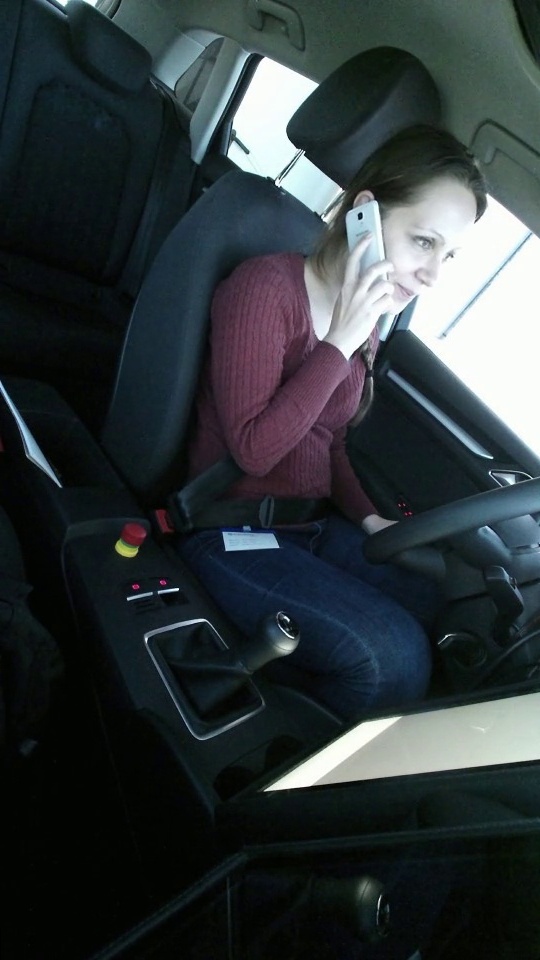}\includegraphics[width=0.11\textwidth,height=0.12\textwidth]{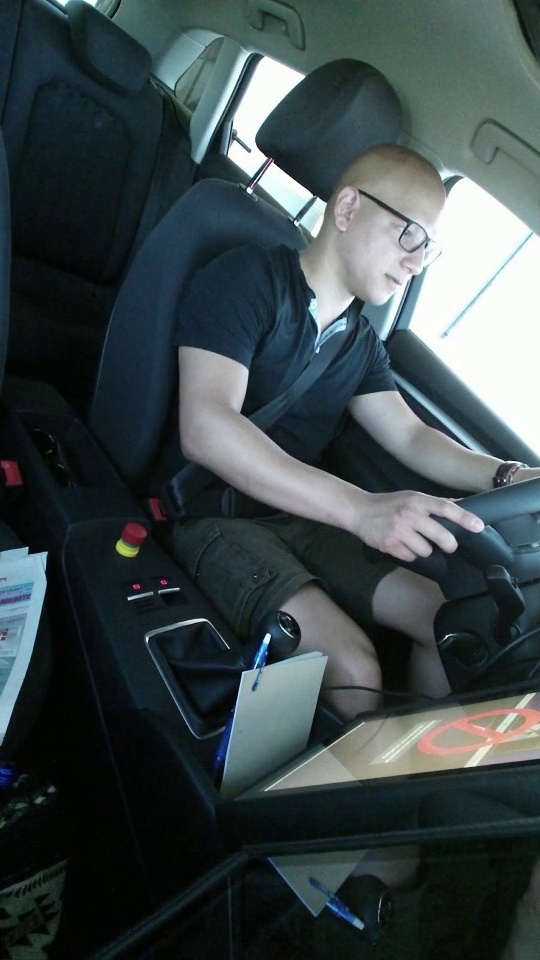}\includegraphics[width=0.1\textwidth,height=0.12\textwidth]{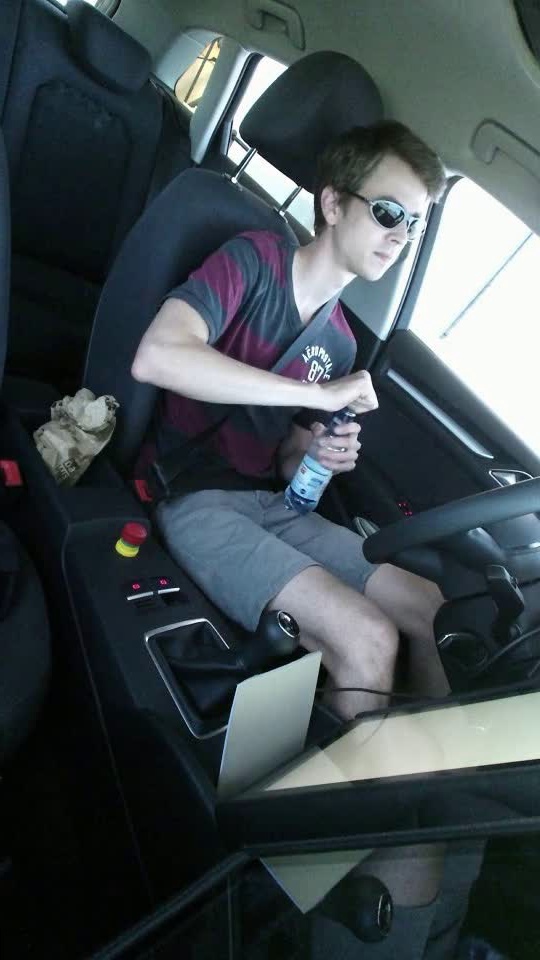}\label{fig:dataset1}}
    \hfill
    %\subfloat[Drive\&Act \cite{martin2019drive}]{\includegraphics[angle=-90,totalheight=0.15\textwidth]{eccv2020kit/Fig/drive_act_v8.jpg}\includegraphics[angle=-90,totalheight=0.15\textwidth]{eccv2020kit/Fig/drive_act_vp14.jpg}\includegraphics[angle=-90,totalheight=0.15\textwidth]{eccv2020kit/Fig/drive_act_3.jpg}\label{fig:dataset1}}
    %\hfill https://www.overleaf.com/project/5e3434e86a5fb60001d63539
    \subfloat[Distracted Driver V1 \cite{abouelnaga2017real}]{\includegraphics[width=0.11\textwidth,height=0.12\textwidth]{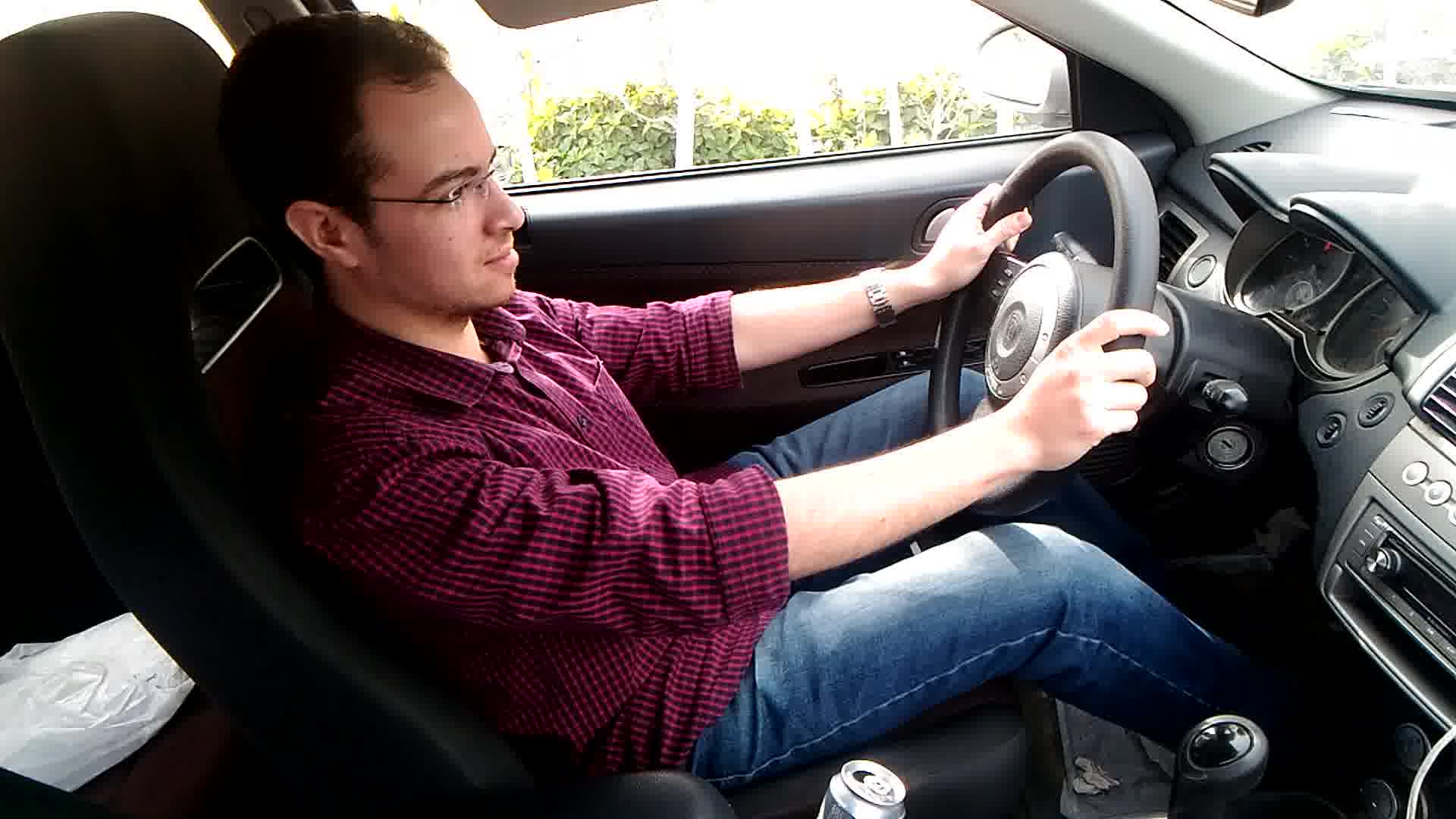}\includegraphics[width=0.11\textwidth,height=0.12\textwidth]{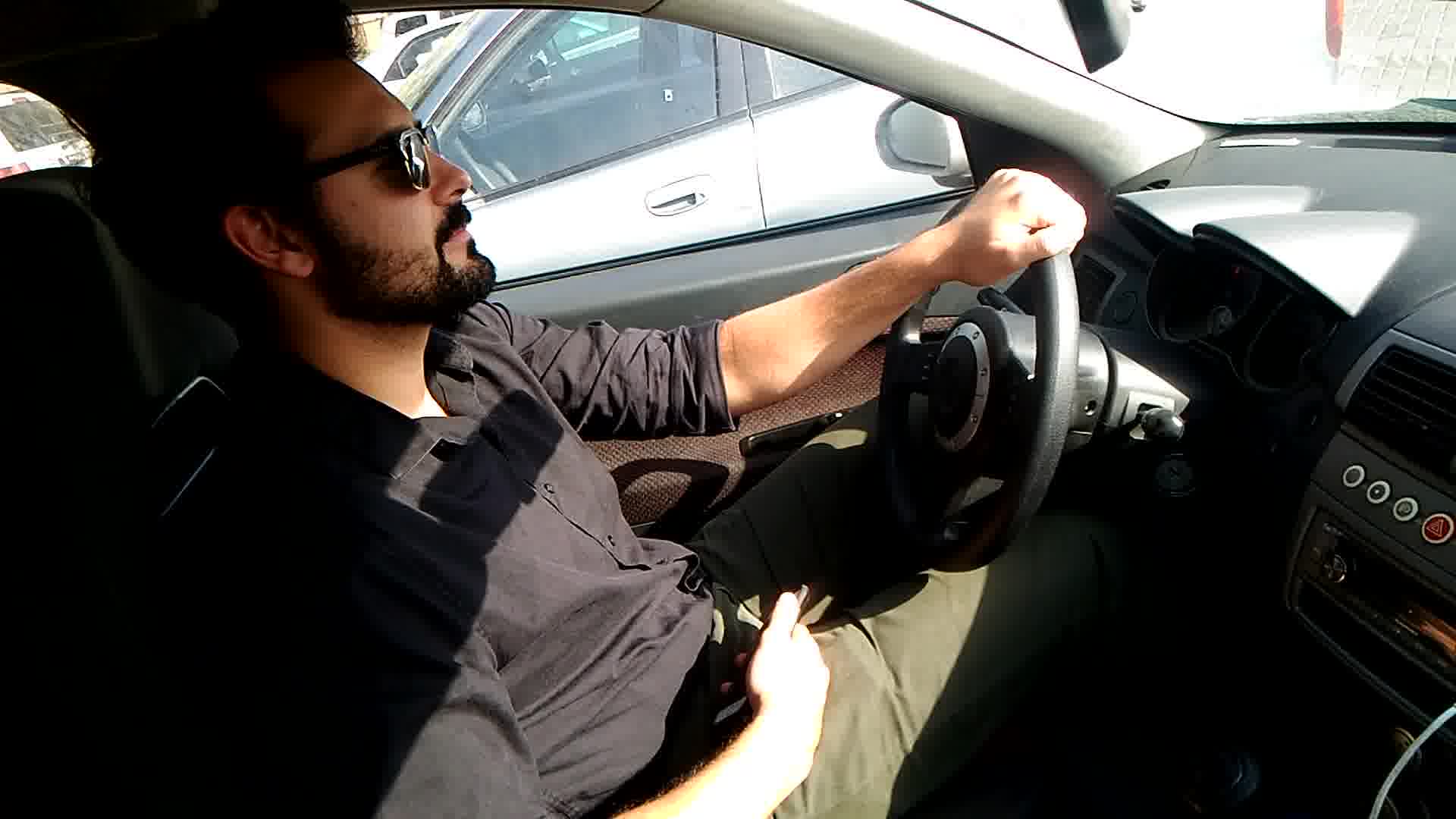}\includegraphics[width=0.11\textwidth,height=0.12\textwidth]{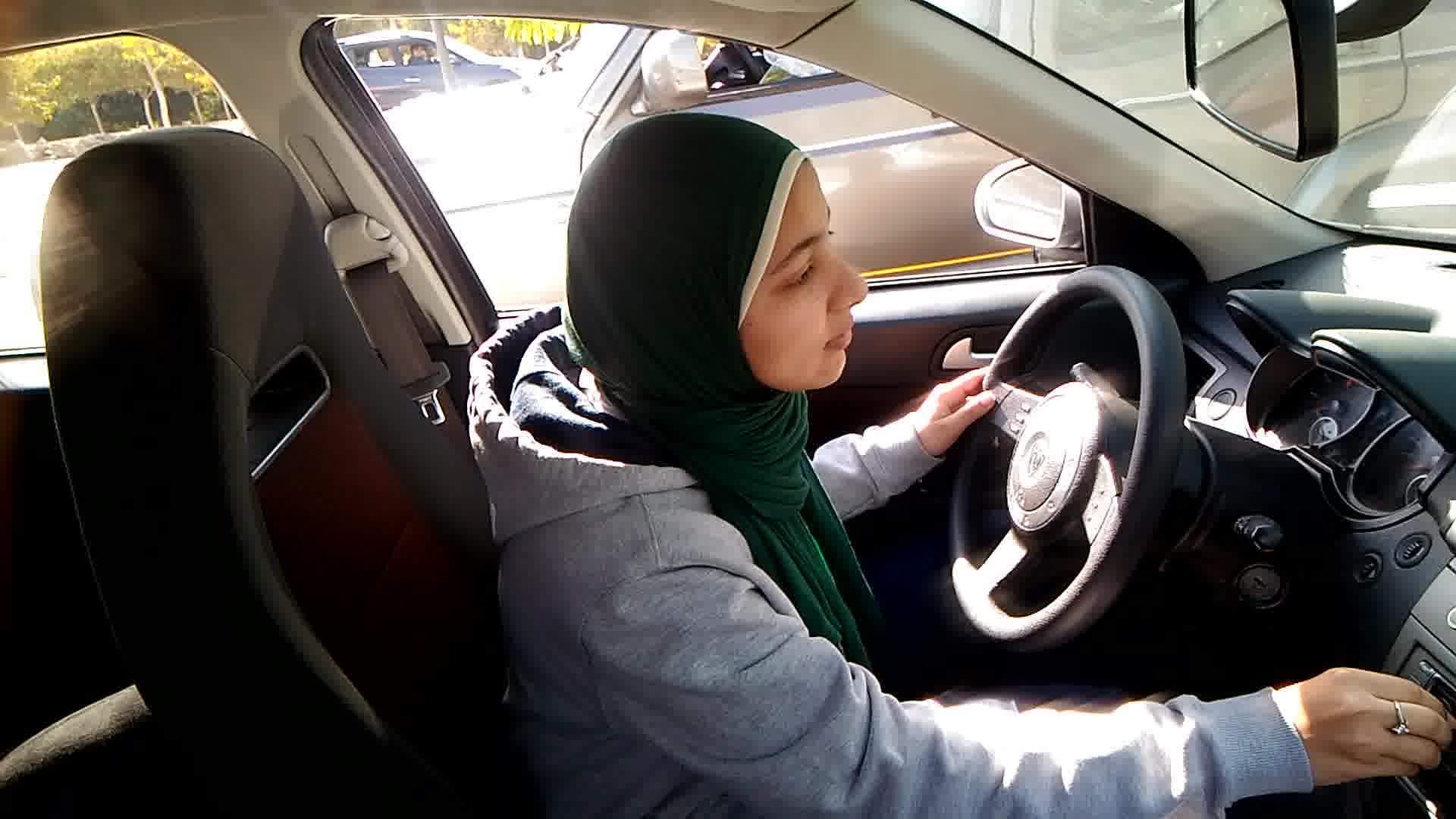}\label{fig:dataset2}}
    \hfill
    \subfloat[Distracted Driver V2
    \cite{eraqi2019driver}]{\includegraphics[width=0.11\textwidth,height=0.12\textwidth]{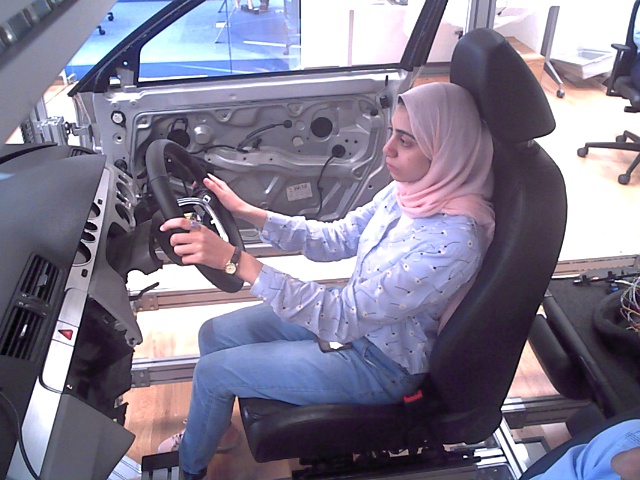}\includegraphics[width=0.11\textwidth,height=0.12\textwidth]{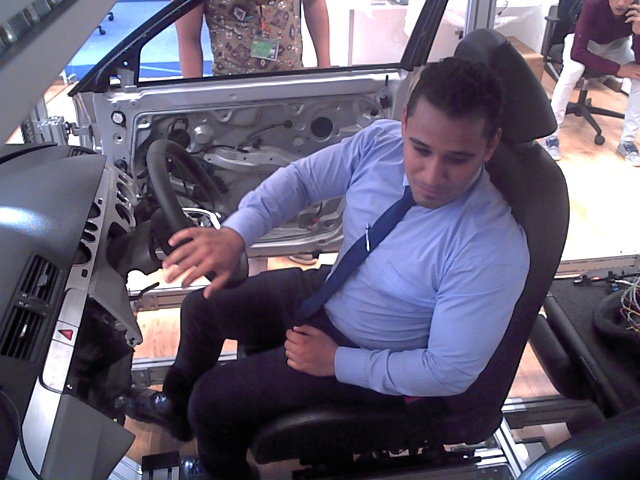}\includegraphics[width=0.11\textwidth,height=0.12\textwidth]{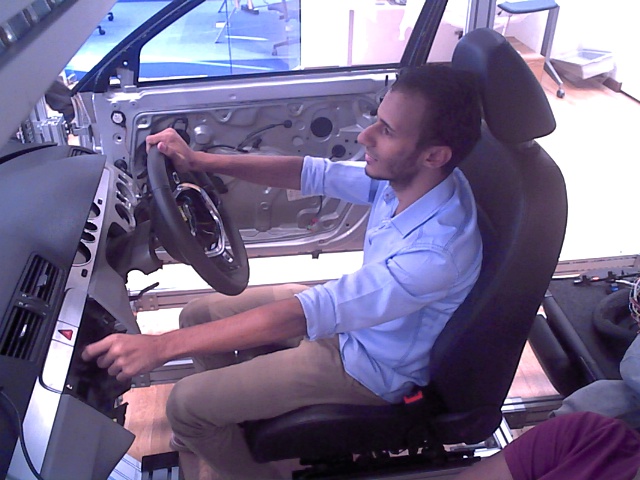}\label{fig:dataset3}}
    \caption{Examples from the datasets used to evaluate our model.}
    \label{fig:dataset}
    \vspace{-0.3cm}
\end{figure*}
\subsection{Temporal attention architecture}\label{sec:temp_attn}
The temporal attention sub-module receives a sequence of glimpse vector $x = (x_1, x_2, \hdots ,x_{T})$. The goal is to encode $x$ using an internal state that summarizes information extracted from the history of past observations. Such state encodes the sequence knowledge and is instrumental in deciding how to act. A common approach to model this state is to use hidden units $h_t\in \mathbb{R}^n$ of the recurrent network and is updated over time as: $h_t=f_h(h_{t-1}, x_t;\theta_h)$, where $f_h$ is a nonlinear function with parameter $\theta_h$. %We choose a fully-gated Long Short-Term Memory (LSTM) network as $f_h$ \cite{hochreiter1997long} and $\theta_h$ is its learnable parameters involving weight matrices and biases linking input, forget and output gates and a cell state. For simplicity, we omitted these parameters from the equation, and we refer the reader to \cite{hochreiter1997long} for further details. For a better understanding of temporal dependency, we show the unrolled LSTM in Fig \ref{fig:figC} but actually, it is the same LSTM. The LSTM layer generates output as a sequence of hidden states $h=(h_1,h_2,\dots h_T)$ corresponding to the input sequence of glimpse vector $x = (x_1,x_2,\dots x_T)$. 
It provides a prediction at each time step $t$, and the sequence recognition is generally carried out by considering prediction in the last time step $T$ based on the associated feature and the previous context vector involving hidden states. This is an inherent flaw in LSTM since the model uses recurrent connections to maintain and communicate temporal information. Therefore, researchers have recently explored temporal pooling (e.g. sum, average, etc.) \cite{sharma2015action} and temporal attention for dynamical pooling \cite{yeung2018every} as additional direct pathways for referencing previously seen frames. Our temporal attention is inspired by \cite{yeung2018every} and focuses on only hidden states of the LSTM. The novelty is to allow the model \textit{learns to attend} automatically the different parts of the hidden states $h$ at each step of the output generation. We achieve this by introducing an attention-focused weighted summation $s = f_a(a_t, h_t;\theta_a)$, where $\theta_a$ consists of learnable weight matrices and biases to compute the attention-focused hidden state representation $a_t$ at $t$. 
\begin{equation}
%\begin{split}
    a_t=h_t + \sum_{t'= 1}^T \beta_{t,t'}h_{t^{'}}, \text{where } \beta_{t,t'} = %\frac{\text{exp}(g_{t,t'})}{\sum_{t'=1}^{T}\text{exp}(W_g \psi_{t,t'}+ b_g)}, 
    \sigma(W_g \psi_{t,t'}+ b_g)%\\
    %g_{t,t'} & = W_g \psi_{t,t'}+ b_g, \text{ and }
    %\psi_{t,t'}&=tanh(W_\psi h_t+W_{\psi^{'}} h_{t^{'}} + b_\psi)
%\end{split}
\vspace{-.2cm}
\end{equation}
\begin{center}
    $\psi_{t,t'}=tanh(W_\psi h_t+W_{\psi^{'}} h_{t^{'}} + b_\psi)$
\end{center}
The element $a_t$ is computed as a residual connection of hidden state representations $h_t$ of the input feature $x_t$ at time $t$. %, the weighted summation of hidden state representations $h_{t^{'}}$ of other input features $x_{t'}$ at time $t'$, and their similarity $\beta_{t,t'}$. % to the hidden state $h_t$ of the current glimpse feature.
The similarity map $\beta_{t,t'}$ is computed from $\psi_{t,t'}$ using the element-wise sigmoid function $\sigma$ and capturing the similarity between the LSTM's hidden state responses $h_t$ and $h_{t^{'}}$. Basically, $a_t$ dictates how much to \textit{attend} the LSTM's current response \textit{conditioned on their neighborhood contexts}. %The element $\beta_{t,t'}$ captures the similarity between the LSTM's hidden state responses $h_t$ and $h_{t^{'}}$ to glimpse features $x_t$ and $x_{t^{'}}$ at the respective time steps $t$ and $t'$. 
$W_\psi$ and $W_{\psi^{'}}$ are the weight matrices for the corresponding hidden states $h_t$ and $h_{t^{'}}$; $W_g$ is the weight matrix for their nonlinear combination; $b_\psi$ and $b_g$ are the bias vectors.

The sequence of attention-focused residual activation $\mathcal{A}=(a_1,a_2,\hdots ,a_T)$ is then used to compute the activity probability as shown in Fig \ref{fig:figC}. We achieve this by using a simple approach of weighted summation: % \cite{felbo2017using} with learnable parameters $W_\phi$ and $b_\phi$.
\begin{equation}
  s = \sum_{t=1}^{T} w_t a_t, \text{ where } w_t = \frac{\text{exp}(a_tW_\phi + b_\phi)}{\sum_{t=1}^{T}\text{exp}(a_tW_\phi + b_\phi)}%, \phi_t = a_tW_\phi + b_\phi   
\end{equation}
Here, $w_t$ provides the score (probability) for each attention-focused residual activation $a_t$ and is computed using weight $W_\phi$ and bias $b_\phi$. Finally, the weighted summation $s$ is then used by a \texttt{Softmax} to estimate the activity probability of a given input video. The parameter $\theta_a=\{W_\psi , W_{\psi^{'}}, W_g , W_\phi , b_\psi, b_g , b_\phi \}$ is learned during training. % of the network.
\subsection{Training} \label{sec:train}
The parameter $\theta=\{\theta_g, \theta_h, \theta_a\}$ of our model consists of %three parameters which are the 
glimpse $\theta_g$, LSTM network $\theta_h$, and the temporal attention network $\theta_a$. The glimpse $f_g$ is implemented with the ResNet-50 \cite{He16} (Fig. \ref{fig:figA}, Section \ref{sec:glimpse}), and initialized with ImageNet's pre-trained weights. %from  ResNet-50 model trained on ILSVRC 2012 ImageNet dataset \cite{russakovsky2015imagenet} except the self-attention layers. The learnable weight matrices in $\theta_a$ (temporal attention) and in $\theta_g$ (glimpse sensor) are initialized with `glorot normal' and `glorot uniform' \cite{glorot2010understanding}, respectively. All the learnable bias vectors are initialized with zeros. 
We use the standard implementation of fully-gated LSTM network $f_h$ \cite{hochreiter1997long} with parameter $\theta_h$. %  including input, forget and output gates and cell states. The  of the LSTM consist of weight matrices and bias vectors, and use the default initialization. %Similarly, the temporal attention network $\theta_a$ also comprises of weight matrices and bias vectors (refer section \ref{sec:temp_attn}), and are randomly initialized. 
These are learned via end-to-end training. 
%Following \cite{behera2018context}, 

We uniformly sample 12 frames from each video segment. The frames are resized to $224\times 224$, and %then fed into the glimpse. %Given an activity video segment of 3 seconds or less, our goal is to assign the correct activity label. 
we use the standard evaluation metric of %average per class accuracy by using the mean of the top-1 accuracy rate of every activity category. 
the top-1 accuracy. % to compare with the state-of-the-art. 
Our model is trained using the Adam optimizer \cite{kingma2014adam} with an initial learning rate of 0.001, and parameters $\beta_1 = 0.9$ and $\beta_2 = 0.999$. The learning rate is reduced by a factor of $0.1$ after every 25 epochs. The experiments are performed on an Ubuntu PC with an Intel Core i9 9820X CPU and a Titan V GPU (12 GB). A batch size of 4 videos is used. % to fit in the 12GB GPU. The inference time of our model is $\sim$1ms per video.
%Our approach is implemented in Keras with TensorFlow as a backend.
%
%
%\vspace{-0.1cm}
%
%------------------------------------------------------------------------
\section{Experimental Results} \label{sec:exp}
\subsection{Datasets and evaluation metric}\label{sec:dataset}
We evaluate our model on three popular driving datasets: 1) Drive\&Act \cite{martin2019drive}, 2) Distracted Driver V1 \cite{abouelnaga2017real}, and 3) Distracted Driver V2 \cite{eraqi2019driver}. To the best of our knowledge, these are only available video datasets for secondary driving activity recognition %. Example images from these datasets are shown in 
(Fig. \ref{fig:dataset}). We also further evaluate our model using SBU Kinect Interaction \cite{yun2012two} dataset consisting of traditional human activities. %Top-1 accuracy is used as the evaluation metric. %We evaluate 3 datasets containing activities performed by drivers behind the wheel of a car: Distracted Driver V1[], Distracted Driver V2[], and Drive \& Act[]. Both Distracted Driver V1 \& V2 are sequential image datasets that have been converted to action-wise video clips per participant; which is the same format used by Drive \& Act.

\noindent \textbf{Drive\&Act \cite{martin2019drive}:} This is a large-scale video dataset (over 9.6 million frames) consisting of various driver activities. %The dataset comprises 12 hours of video captured by multi-modal synchronized cameras placed in six different positions. There are 15 participants (11 male and 4 female) and each one has a total of around 50 minutes of video footage from 2 separate sessions. %The dataset also Depth data is also available from several angles but only the RGB data is used. 
Annotations are provided for 12 classes (full scene actions) of top-level activities (e.g. eating and drinking), 34 categories (semantic actions) of fine-grained activities (e.g. opening bottle, preparing food, etc.), and 372 classes (object interactions) of \textit{atomic action unit} involving triplet of \textit{action}, \textit{object}, and \textit{location}. There are 5 types of actions, 17 object classes and 14 location annotations. %The dataset is divided randomly into three splits based on the participant identity. For each split, the data from ten subjects is used for training, two subjects for validation, and three subjects for testing (i.e. 20, 4 and 6 driving sessions, respectively). Finally, each action segment is split up into chunks of up to 3-second or less. For example, a 5 second action might result in one 3 second sequence, and another 2-second sequence. 
We follow the same three splits based on the participant identity and use the same train, test, and validation sets in each split as those in \cite{martin2019drive}. Final result is the average over the three splits.

\noindent \textbf{Distracted Driver V1 \cite{abouelnaga2017real}:} It contains 12977 train and 4331 test images from 31 drivers (22 male and 9 female) from 7 different countries. There are 10 activity classes (e.g. safe driving, texting, etc.). % – right, 4) texting - left, 5) talking on phone – left, 6) adjusting radio, 7) drinking, 8) reaching behind, 9) hair \& makeup, and 10) talking to passenger. 
It consists of videos of each subject, but the frame-based evaluation is carried out in \cite{abouelnaga2017real}, subject-wise video-based evaluation is done in \cite{behera2018context}. We follow the evaluation protocol in \cite{behera2018context}, which uses the videos of 22 participants for training and the rest of the videos for testing. %This is due to Behera et al. using passenger-wise splits rather than randomly shuffled images in the original publication. 
%This has resulted in 220 training videos and 90 test videos (10 videos per participant, i.e. one activity video per subject).

\noindent \textbf{Distracted Driver V2 \cite{eraqi2019driver}:} This is a newer iteration of dataset V1 \cite{abouelnaga2017real}, containing 14478 images from 44 drivers (29 male and 15 female) using the same 10 activities. %A total of 11680 images are reused from the previous dataset. 
The dataset is split into % so that no driver that appears in training also appears in testing. We follow their train/test split of 
12555 (36 drivers) training and 1923 (8 drivers) testing images, respectively. The dataset associated approach \cite{eraqi2019driver} has used the frame-wise evaluation. In this work, we are the first one to provide a video-based evaluation. %As this model focuses on temporal information, each driver activity is still converted into a single video clip. 
A total of 360 videos from 36 participants are used for training and the rest of the videos %from 8 subjects 
are used for testing.

\noindent \textbf{SBU Kinect Interaction \cite{yun2012two}:} The dataset is used to justify our model's wider applicability. It consists of
282 videos with 8 different activity classes. It contains interactions between two subjects and is close to the driver's secondary activities involving human-objects and human-car interactions. We follow the same train/test split in \cite{yun2012two}.  
\begin{figure*}[t]
    \centering
    \includegraphics[width=0.87\textwidth]{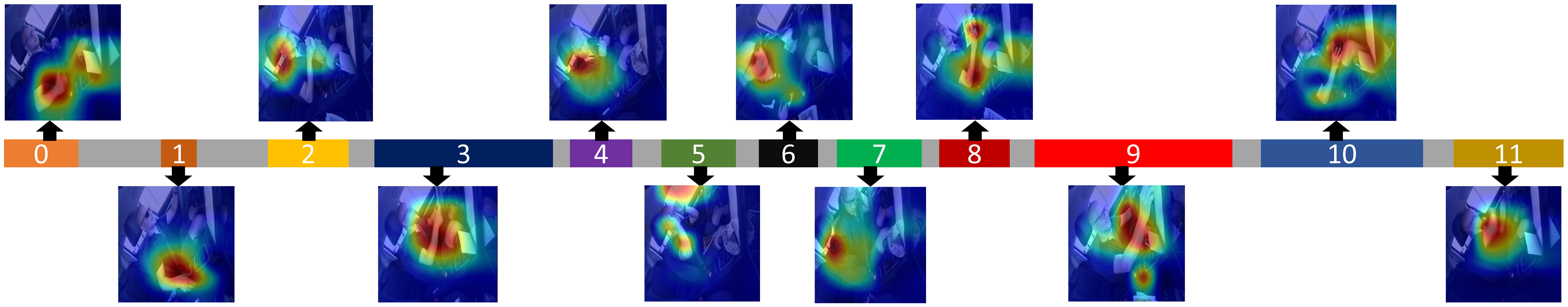}
    \caption{Timeline of a video example from the Drive\&Act dataset displaying 12 different coarse activities executed by a subject. The duration of each activity is represented by the respective color bar. A visual explanation of the classification decision is overlaid using the class activation map \cite{selvarajuvisual} representing salient regions of various activities over the video sequence. Scenarios are: (0) fasten seat belt (and get in vehicle); (1) hand over (turn on autonomous vehicle); (2) eat and drink; (3) read newspaper; (4) put on sunglasses; (5) take off sunglasses; (6) put on jacket; (7) take off jacket; (8) read magazine; (9) watch video (on vehicle display); (10) work (type on laptop); and (11) final task (get out of vehicle). Best view in color.}
    \label{fig:time_line}
\vspace{-0.1cm}
\end{figure*}
\begin{table*}[t]
\begin{center}
        \begin{tabular}{|l|cc||cc||cc||cc||cc||cc|}
        \hline
        \textbf{Model} &\multicolumn{2}{|c||}{\textbf{Fine-grained}} & \multicolumn{2}{c||}{\textbf{Coarse task}} &\multicolumn{2}{c||}{\textbf{Action}} & \multicolumn{2}{c||}{\textbf{Object}} & \multicolumn{2}{c||}{\textbf{Location}} & \multicolumn{2}{c|}{\textbf{All}}  \\
        & Val & Test & Val & Test & Val & Test & Val & Test & Val & Test & Val & Test \\
        %\hline
        %Random & 16.67 & 16.67 & 5.88 & 5.88 & 7.14 & 7.14 & 0.39 & 0.31 \\
        \hline
        Pose \cite{martin2019drive} & 53.17 & 44.36 & 37.18 & 32.96 & 57.62 & 47.74 & 51.45 & 41.72 & 53.31 & 52.64 & 9.18 & 7.07 \\
        Interior \cite{martin2019drive} & 45.23 & 40.30 & 35.76 & 29.75 & 54.23 & 49.03 & 49.90 & 40.73 & 53.76 & 53.33 & 8.76 & 6.85 \\
        2-Stream \cite{wang2017modeling} & 53.76 & 45.39 & 39.37 & 34.81 & 57.86 & 48.83 & 52.72 & 42.79 & 53.99 & 54.73 & 10.31 & 7.11 \\
        3-Stream \cite{martin2018body} & 55.67 & 46.95 & 41.70 & 35.45 & 59.29 & 50.65 & 55.59 & 45.25 & \textbf{59.54} & 56.50 & 11.57 & 8.09 \\
        \hline
        C3D \cite{tran2015learning} & 49.54 & 43.41 &- &- &- &- &- &- &- &- &- &-\\
        P3D Net \cite{qiu2017learning} & 55.04 & 45.32 &- &- &- &- &- &- &- &- &- &-\\
        I3D Net \cite{carreira2017quo}  & 69.57 & 63.64 & 44.66 & 31.80 & \textbf{62.81} & 56.07 & 61.81 & 56.15 & 47.70 & 51.12 & 15.56 & 12.12 \\
        \hline
        \textbf{CTA-Net} & \textbf{72.42} & \textbf{65.25} & \textbf{62.82} & \textbf{52.31} & 57.59 & \textbf{56.41} & \textbf{63.37} & \textbf{59.19} & 56.41 & \textbf{63.01} & \textbf{46.44} & \textbf{49.41} \\
        \hline
    \end{tabular}
\caption{Recognition results (Validation and Testing accuracy in \%) of the fine-grained and coarse tasks, as well as Atomic Action Units defined as \{\textit{Action}, \textit{Object}, \textit{Location}\} triplets, and their combinations in Drive\&Act dataset \cite{martin2019drive}. A total of 34 fine-grained, 12 coarse tasks. There are 5 actions, 17 object categories, 14 locations and 372 (\textbf{All}) possible combinations.}
\label{table:t2}
\end{center}
\vspace{-2em}
\end{table*}
\subsection{Results and comparative studies} \label{sec:experiments}
We first compare the CTA-Net with the state-of-the-art on Drive\&Act dataset. An example of a 12 coarse activity video with a duration of 27 minutes is shown in Fig. \ref{fig:time_line}. In this figure, we have also shown the class activation map \cite{selvarajuvisual} representing the visual explanation of the classification decision of our model for various coarse scenarios. The accuracy (\%) of our model and state-of-the-art approaches for recognizing 12 coarse and 34 fine-grained activities is presented in Table \ref{table:t2}. It is observed that the CTA-Net outperforms in both validation and testing sets by a significantly large margin. For example, in coarse activity, CTA-Net (62.82\%) is 18.2\% higher than the best model (I3D Net \cite{carreira2017quo}: 44.66)  and 16.9\% higher than the three-stream \cite{martin2018body} (35.45\%) on the respective validation and test set. Similarly, I3D Net is the best performer (Val: 69.57\% and Test: 63.64\%) in recognizing fine-grained activities. Our CTA-Net outperforms these by a margin of 2.85\% (Val) and 1.61\% (Test), respectively. It is seen that the margin of improvement in recognizing coarse activities (Val: 18.2\% and Test 16.9\%) is significantly larger than those of fine-grained ones. This suggests that our model can effectively capture long-term dependencies. %in videos and is appropriate to recognize long-range activities. 
This is due to the introduction of novel coarse temporal branches %in the proposed glimpse sensor 
to model the ‘during’, ‘before’, and ‘after’ temporal relationships explicitly in videos. Moreover, I3D Net is developed to recognize distinctive human activities and is used here to recognize the driver's activities involving subtle changes. This suggests that it might not be suitable for such applications. The visual explanation using class activation map \cite{selvarajuvisual} representing our coarse temporal relationships in `reading magazine' and `exiting vehicle' activities is shown in Fig. \ref{fig:attn2} and Fig. \ref{fig:attn3}, respectively. More examples are included in the supplementary.
The confusion matrix using our CTA-Net is shown in Fig \ref{fig:conf_coarse} for the coarse tasks %(12 scenarios) in the test set of 
in the Drive\&Act dataset. %The confusion matrix is generated using average over all three splits. 
It is clear that the performance of activities `watching videos' (class 9), `final task' (class id 11, get out of vehicle),  `take off sunglasses' (class 5) and `turn on AV feature' (class 1) is low. This is mainly due to the involvement of very little action in `watching videos' and `turn on AV feature' activities except pressing a button. Thus, watching a video is confused with `turn on AV feature'. The `take off sunglasses' activity is confused with `put on sunglasses' and `turn on AV feature' since sunglasses is a small object representing very little visual information. Moreover, the sunglasses are kept in the holder close to the vehicle touch screen, confusing with `turn on AV feature'. Similarly, `getting in' is confused with `getting out' since there are no significant visual changes but, motion direction information would help in discriminating such activities. The confusion matrix for the fine-grained activities %(34 classes) is presented in Fig. \ref{fig:conf_fine}. We have also included 
and the split-wise confusion matrices of both coarse and fine-grained activities are included in the supplementary material.

The accuracy of the Atomic Action Units \{\textit{Action}, \textit{Object}, \textit{Location}\} is provided in Table \ref{table:t2}. Like in coarse and fine-grained activities, the CTA-Net outperforms in each triplet, as well as their unique 372 combinations %. There are 372 possible combinations when the action, object, and location are combined 
(\textbf{All} in Table \ref{table:t2}). A notable performance of our model can be seen for recognizing the above combinations. The best performer is 15.56\% (Val) and 12.12\% (Test) by the I3D Net \cite{carreira2017quo}. Whereas, the proposed approach is significantly better (Val: 46.44\% and Test: 49.41\%). This is mainly due to our self-attention module (Fig. \ref{fig:figB}), which explicitly learns the relationships between pixels located at the \textit{CONV4} output (Fig. \ref{fig:figA}). It allows to capture the subtle changes within a video frame to discriminate the unique combinations of \textit{action}-\textit{object}-\textit{location}. This suggests that our model is not only suitable for recognizing long-term dependencies in videos, but also appropriate in classifying atomic action units involving action, location, objects and their distinct combinations. This is due to the design, which considers both coarse temporal attention to model high-level temporal dependencies (glimpse in Section \ref{sec:glimpse}) and fine-grained temporal attention for each frame by weighing them (Section \ref{sec:temp_attn}) when constructing the representation of an input video. The proposed approach also performs better than the state-of-the-art for individual atomic action units except in \textit{location} and \textit{action} validation sets. For \textit{location}, our accuracy (56.41\%) is not far from the best (59.54\%) \cite{martin2018body} that combines three streams, whereas our approach uses only the RGB video stream. For \textit{action}, I3D Net \cite{carreira2017quo} performed (62.81\%) better %higher than ours (57.59\%) 
in the validation set, but in the testing set, ours is slightly better. This could be due to the \textit{action} consisting of atomic verbs such as opening, closing, reaching for, etc. These are very minimal duration and %do not link to any objects. Therefore, 
thus, inflated 2D convolution is appropriate in capturing 3D spatiotemporal information resulting in higher accuracy.  

%-------------------------------------------------------------------------
Table \ref{table:t4} presents our CTA-Net's accuracy on Distracted Driver V1 \cite{abouelnaga2017real} and V2 \cite{eraqi2019driver} datasets. Both datasets consist of the video sequence. The existing approaches use frame-wise evaluation on V2 \cite{eraqi2019driver}, and we are the first one to provide a video-based evaluation. % on this dataset. %, but the frame-wise evaluation is used. 
The Multi-stream LSTM \cite{behera2018context} has used the video-based evaluation on V1 \cite{abouelnaga2017real} and we followed it to evaluate our CTA-Net. In \cite{behera2018context}, multiple streams focusing on body pose and body-object interactions, and CNN features are used by an LSTM to recognize various activities, whereas we only focus on RGB video. The accuracy of our approach is significantly (84.09\%) better. % than the recent best \cite{behera2018context} (52.22\%). 
Similarly, the accuracy of our model is 92.5\% on V2 \cite{eraqi2019driver}. 
\begin{figure*}[t]
\begin{center}
\begin{minipage}[b]{.27\textwidth}
  \subfloat
    [Confusion matrix (Coarse tasks)]
    {\label{fig:conf_coarse}\includegraphics[width=\textwidth]{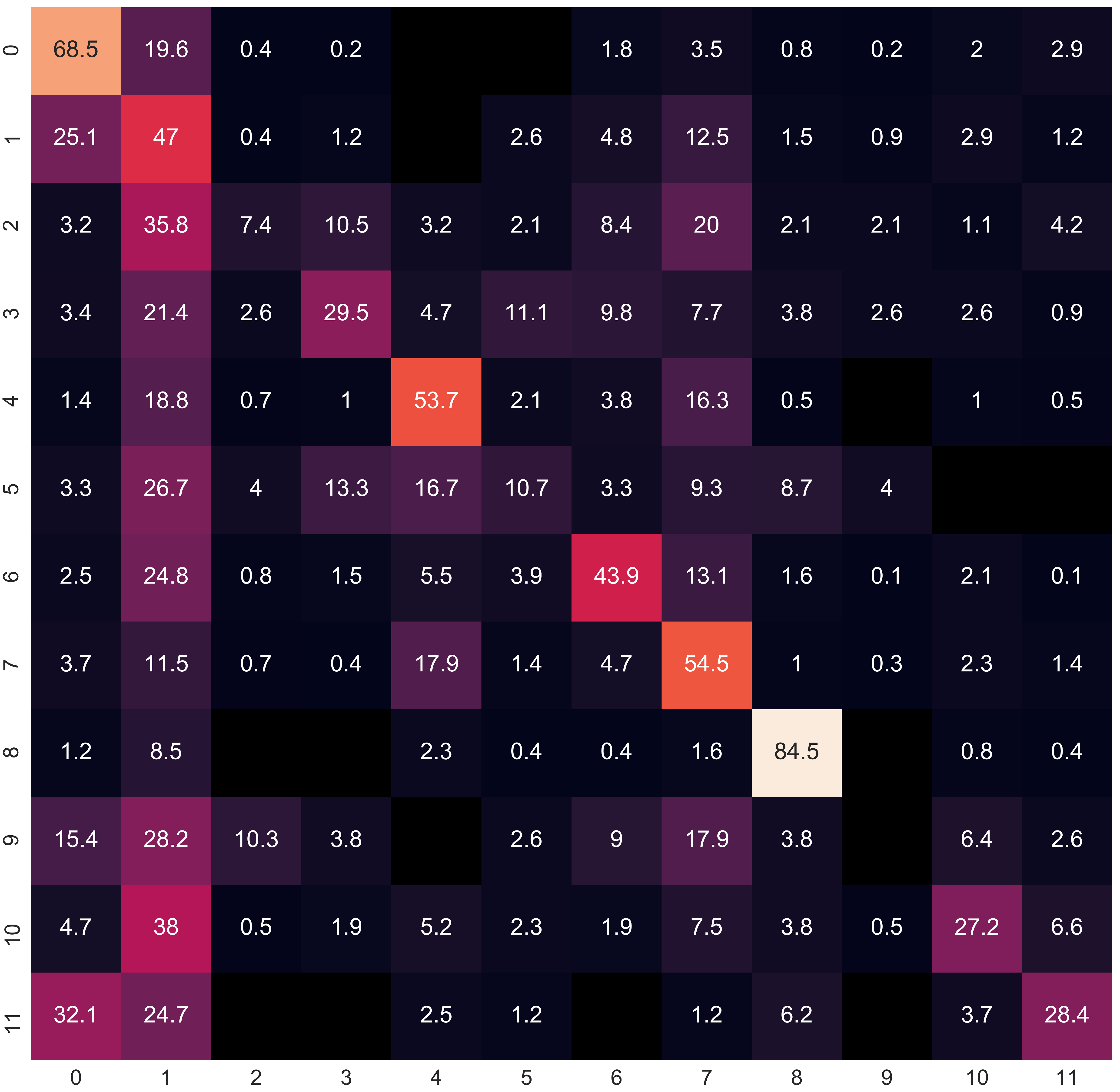}}
  \end{minipage}
\begin{minipage}[b]{.55\textwidth}
\centering
\subfloat[Reading: `before', `during', and `after']
  {\label{fig:attn2}\includegraphics[width=0.2\textwidth]{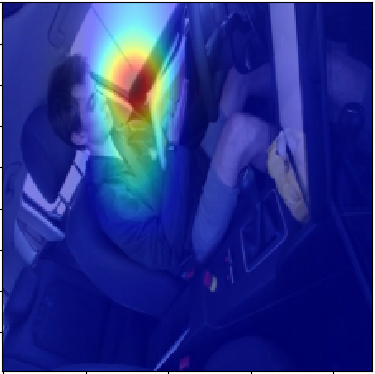}\hspace{.25cm} \includegraphics[width=0.2\textwidth]{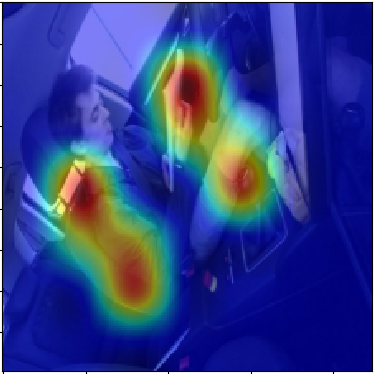}\hspace{.25cm} \includegraphics[width=0.2\textwidth]{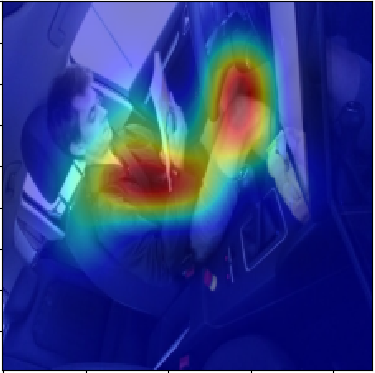}}  
\vfill
\subfloat[Exiting car: `before', `during', and `after']
  {\label{fig:attn3}\includegraphics[width=0.2\textwidth]{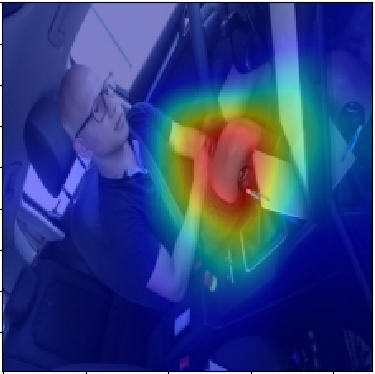}\hspace{.25cm} \includegraphics[width=0.2\textwidth]{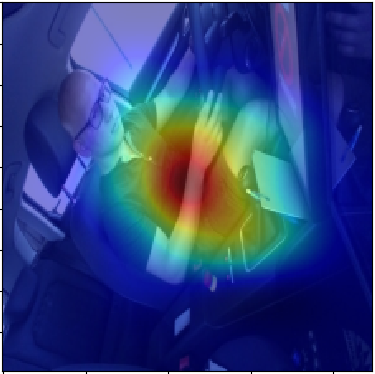}\hspace{.25cm} \includegraphics[width=0.2\textwidth]{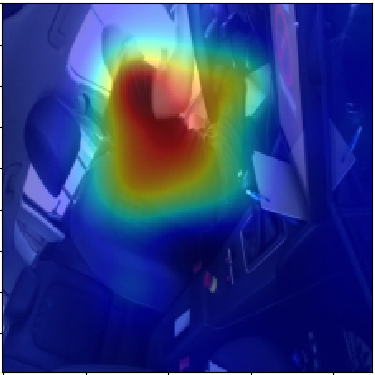}} 
\end{minipage}
\caption{a) Our CTA-Net's confusion matrix showing 12 coarse tasks in the Drive\&Act test set. A visual explanation of decision using class activation map \cite{selvarajuvisual} representing our coarse temporal attention of `before' (left), `during' (middle), and `after' (right) segment of an input video with b) reading activity and c) exiting the vehicle. Best view in color.}
\end{center}
\vspace{-0.5cm}
\end{figure*}
\begin{table}[t]
\begin{center}
\begin{tabular}{|l|c||c|c|}
\hline\noalign{}
\multicolumn{2}{|c||}{\textbf{Distracted Driver V1 \cite{abouelnaga2017real}}} & \multicolumn{2}{c|}{\textbf{Distracted Driver V2 \cite{eraqi2019driver}}}\\
\hline\noalign{}
\textbf{Model} & \textbf{ACC} &\textbf{Model} & \textbf{ACC}\\
\hline\noalign{}
One-stream \cite{behera2018context} & 42.22 &Incep. V3$^*$ \cite{szegedy2016rethinking} &90.07\\
Two-streams \cite{behera2018context} & 44.44 &ResNet-50$^*$ \cite{He16} & 81.70\\
Three-streams \cite{behera2018context} &52.22 &VGG-16$^*$ \cite{simonyan2014very} &76.13\\
Four-streams \cite{behera2018context} &37.78 & &\\
\hline\noalign{}
\textbf{CTA-Net} &\textbf{84.09} & \textbf{CTA-Net} &\textbf{92.50}\\
\hline\noalign{}
\end{tabular}
\caption{Recognition accuracy (\%) of 10 different driver's activities using Distracted Driver datasets. $*$ These methods are used for frame-wise evaluation.}
\label{table:t4}
\end{center}
\vspace{-0.8cm}
\end{table}
\begin{table}[t]
    \centering
    %\vspace{0.2mm}
    \begin{tabular}{|l|c|c|c|c|}
    \hline
         \textbf{Approaches} & \textbf{Pose} & \textbf{RGB} & \textbf{Depth} & \textbf{ACC}\\
         \hline
         Raw Skeleton \cite{yun2012two} & \checkmark & - & - & 49.7\\
         Joint Feature \cite{yun2012two} & \checkmark & - & - & 80.3\\
         Raw Skeleton \cite{ji2014interactive} & \checkmark & - & - & 79.4\\
         Joint Feature \cite{ji2014interactive} & \checkmark & - & - & 86.9\\
         Co-occ. RNN \cite{zhu2016co} & \checkmark & - & - & 90.4\\
         STA-LSTM \cite{song2017end} & \checkmark & - & - & 91.5\\
         ST-LSTM \cite{liu2016spatio} & \checkmark & - & - & 93.3\\
         DSPM \cite{lin2016deep} & - & \checkmark & \checkmark & 93.4\\
         Ijjina \cite{ijjina2017human} & \checkmark & - & - & 82.2 \\
         Ijjina \cite{ijjina2017human}& - & \checkmark & \checkmark & 85.1 \\
         Baradel \cite{baradel2018human} & \checkmark & - & - & 90.5\\
         Baradel \cite{baradel2018human} & \checkmark & \checkmark & - & 94.1\\
         \hline
         Ijjina \cite{ijjina2017human} & - & \checkmark & - & 75.5 \\
         Baradel \cite{baradel2018human} & - & \checkmark & - & 72.0\\
         \hline
         \textbf{CTA-Net} & - & \checkmark & - & \textbf{92.9}\\
         \hline\noalign{} 
    \end{tabular}
    \caption{CTA-Net's accuracy (\%) and its comparison to the state-of-the-art using SBU Kinect Interaction dataset \cite{yun2012two}.}
    \label{table:t5}
    \vspace{-1em}
\end{table}
\begin{table*}[t]
\begin{center}
\begin{tabular}{|c|c|cc|cc|cc|cc|}
        \hline
        \textbf{Annotation} &\textbf{Split} & \multicolumn{4}{c|}{\textbf{Without} \textit{during}, \textit{before} and \textit{after}} & \multicolumn{4}{c|}{\textbf{With} \textit{during}, \textit{before} and \textit{after}}  \\
        \cline{3-10}
        &  & \multicolumn{2}{c|}{\textbf{No Attention}} & \multicolumn{2}{c|}{\textbf{Attention}} & \multicolumn{2}{c|}{\textbf{No Attention}} & \multicolumn{2}{c|}{\textbf{Attention}} \\
        &  & Val & Test & Val & Test & Val & Test & Val & Test \\
        \hline
        \multirow{4}{*}{Fine-grained} & 0 &56.05 &52.35 &51.71 &53.76 &50.36 &44.74 &\textbf{76.97} &\textbf{71.43}\\
        & 1 &49.71 &39.41 &50.59 &45.07 &48.82 &41.50 &\textbf{72.94} &\textbf{67.94} \\
        & 2 &55.30 &43.67 &56.41 &43.98 &53.75 &38.07 &\textbf{67.34} &\textbf{56.85} \\
        %\hline
        & Avg &53.69 &45.14 &52.90 &47.60 &50.98 &41.44 &\textbf{72.42} &\textbf{65.41} \\
        \hline
        \multirow{4}{*}{Coarse scenarios} & 0 &47.41 &43.92 &46.55 &39.80 &43.34 &44.29 &\textbf{63.09} &\textbf{61.13}\\
        & 1 &41.94 &44.43 &41.12 &44.91 &38.77 &49.39 &\textbf{55.34} &\textbf{54.34} \\
        & 2 &53.66 &31.23 &60.28 &33.60 &45.73 &30.05 &\textbf{70.02} &\textbf{41.47} \\
        %\hline
        & Avg &47.67 &39.86 &49.32 &39.44 &42.61 &41.24 &\textbf{62.82} &\textbf{52.31} \\
        \hline
\end{tabular}
\end{center}
\vspace{-0.3cm}
\caption{Split-wise accuracy (\%) of fine-grained and coarse scenario activities with and without temporal relationships  (`before', `during', and `after'), as well as with and without our novel attention mechanism using Drive\&Act dataset \cite{martin2019drive}.}
\label{table:t6}
\vspace{-0.1cm}
\end{table*}
\begin{table*}[!h]
\begin{center}
\begin{tabular}{|c||cc||cc||cc||cc||cc||cc|}
        \hline
        \textbf{Split} & \multicolumn{2}{c||}{\textbf{Fine-Grained}} & \multicolumn{2}{c||}{\textbf{Coarse}} &  \multicolumn{2}{c||}{\textbf{Action}} & \multicolumn{2}{c||}{\textbf{Object}} & \multicolumn{2}{c||}{\textbf{Location}} & \multicolumn{2}{c|}{\textbf{All}}  \\
        & Val & Test & Val & Test & Val & Test & Val & Test & Val & Test & Val & Test \\
\hline
0 & 76.97 & 71.43 & 63.09 & 61.13 & 57.82 & 60.94 & 63.01 &57.94 &46.50 & 57.01 & 42.95 & 52.07 \\
\hline 
1 & 72.94 & 67.94 & 55.34 & 54.34 &56.74 &54.88 &62.87 &64.86 & 68.78 & 64.10 & 52.79 & 49.89 \\
\hline 
2 & 67.34 & 56.85 & 70.02 & 41.47 & 58.20 & 53.40 &64.23 &54.77 &53.94 &67.92 & 43.57 & 46.27 \\
\hline 
Avg &72.42 &65.25 &62.82 &52.31 &57.59 &56.41 &63.37 &59.19 &56.41 &63.01 &46.44 &49.41\\
\hline
\end{tabular}
\end{center}
\vspace{-0.2cm}
\caption{Split-wise accuracy (\%) of fine-grained, coarse activities and atomic action units using our model on Drive\&Act. % \cite{martin2019drive}.
}
\label{table:t3}
\vspace{-0.2cm}
\end{table*}

On the SBU Kinect dataset \cite{yun2012two}, our model significantly outperforms (92.9\%) the state-of-the-arts using RGB only (72\% \cite{baradel2018human}, 75.5\% \cite{ijjina2017human}), as shown in Table \ref{table:t5}. Moreover, the accuracy is close to the existing approaches that use multi-modal (RGB+Depth: 93.4\% \cite{lin2016deep}, RGB+Pose: 94.1\% \cite{baradel2018human}) and even better than the approach in \cite{ijjina2017human}, which uses RGB+Depth (85.1\%). However, such multi-modal information is not always available or requires additional devices for data capture. This demonstrates that our CTA-Net is not only suitable for recognizing driver’s activity but also appropriate in classifying traditional human activities. %involving human-objects and human-human interactions with an advantage of easy implementation. 
%
%\vspace{-.2cm}
\subsection{Ablation studies}
We have conducted ablation studies to understand the impact of the proposed high-level temporal relationships (`before', `during', and `after'), as well as our novel attention mechanism (see Section \ref{sec:temp_attn}) on the performance of our model using individual split. The results are shown in Table \ref{table:t6}. It is evident that the performance of combined high-level temporal relationships and attention mechanism is significantly higher than the rest of the combinations. Moreover, the average accuracy (fine-grained: Val 72.42\%, Test 65.41\% and scenario: Val 62.82\%, Test 52.31\%) using `before', `during', and `after' relationships is considerably higher than without them (fine-grained: Val 52.9\%, Test 47.6\% and coarse: Val 49.32\%, Test 39.44\%). This justifies the inclusion of the proposed coarse temporal relationships. Similarly, the performance is higher with the inclusion of our attention mechanism  than without it. This vindicates the significance of the proposed attention mechanism in our model.

We have also provided our model's accuracy using individual split in Drive\&Act (Table \ref{table:t3}). There is not any significant difference in accuracy among the splits, suggesting the splits are balanced. We have also included additional confusion matrices in the supplementary document.  
%We have also carried out ablation study studied the influence of the number (32, 64 and 128) of hidden states in the LSTM layer on fine-grained activities using split 1. It is observed that the test accuracy increases as the numbers increases - 32 (Val: 72.97, Test: 52.90), 64 (Val: 72.97, Test: 54.36) and 128 (Val: 72.94, Test: 67.49). However, there is no improvement in validation accuracy. This could be due to the number of validation samples is less than the test samples. We are unable to experiment with more than 128 states due to GPU limitation of 12GB.  
%
%------------------------------------------------------------------------
%
%To evaluate the wider applicability of the proposed approach, we evaluate it on the well-known SBU Kinect interaction dataset \cite{yun2012two}. This dataset consists of interactions between two subjects and is close to the driver's activities involving human-objects and human-car interactions.  Our model significantly outperforms (Accuracy: 92.9\%)  the state-of-the-art approaches using RGB only (Baradel et al. \cite{baradel2018human}: 72\%, Ijjina et al. \cite{ijjina2017human}: 75.5\%). Moreover, the performance is close to the existing approaches that use multi-modal (RGB+Depth - Lin et al. \cite{lin2016deep}: 93.4\%, RGB+Pose - Baradel et al. \cite{baradel2018human}: 94.1\%) information. We have included the comparison table in the supplementary document. 
\vspace{-0.1cm}
\section{Conclusion}\label{sec:con}
In this paper, we have proposed a novel end-to-end network (CTA-Net) for driver's activity recognition and monitoring by employing an innovative attention mechanism. The proposed attention generates a high-dimensional contextual feature encoding for activity recognition by learning to decide the importance of hidden states of an LSTM that takes inputs from a learnable glimpse sensor. We have shown that capturing coarse temporal relationships (`before', `during', and `after') via focusing certain segments of videos and learning meaningful temporal and spatial changes have a significant impact on the recognition accuracy. Our proposed architecture has notably outperformed existing methods and obtains state-of-the-art accuracy on four major publicly accessible datasets: Drive\&Act, Distracted Driver V1, Distracted Driver V2, and SBU Kinect Interaction. We have demonstrated that the proposed end-to-end network is not only suitable for monitoring driver's activities but also applicable to traditional human activity recognition problems. Finally, our model’s state-of-the-art results on benchmarked datasets and ablation studies justify the design of our approach. Future work will be to apply the proposed technique for the development of the driving assistance system.

\noindent\textbf{Acknowledgements:} This research was supported by the UKIERI (CHARM) under grant DST UKIERI-2018-19-10. The GPU is kindly donated by the NVIDIA Corporation. 

%insert supplementary document contents
%\Large{Supplementary Information}
\begin{figure*}[]
    \centering
    \includegraphics[width=0.98\textwidth]{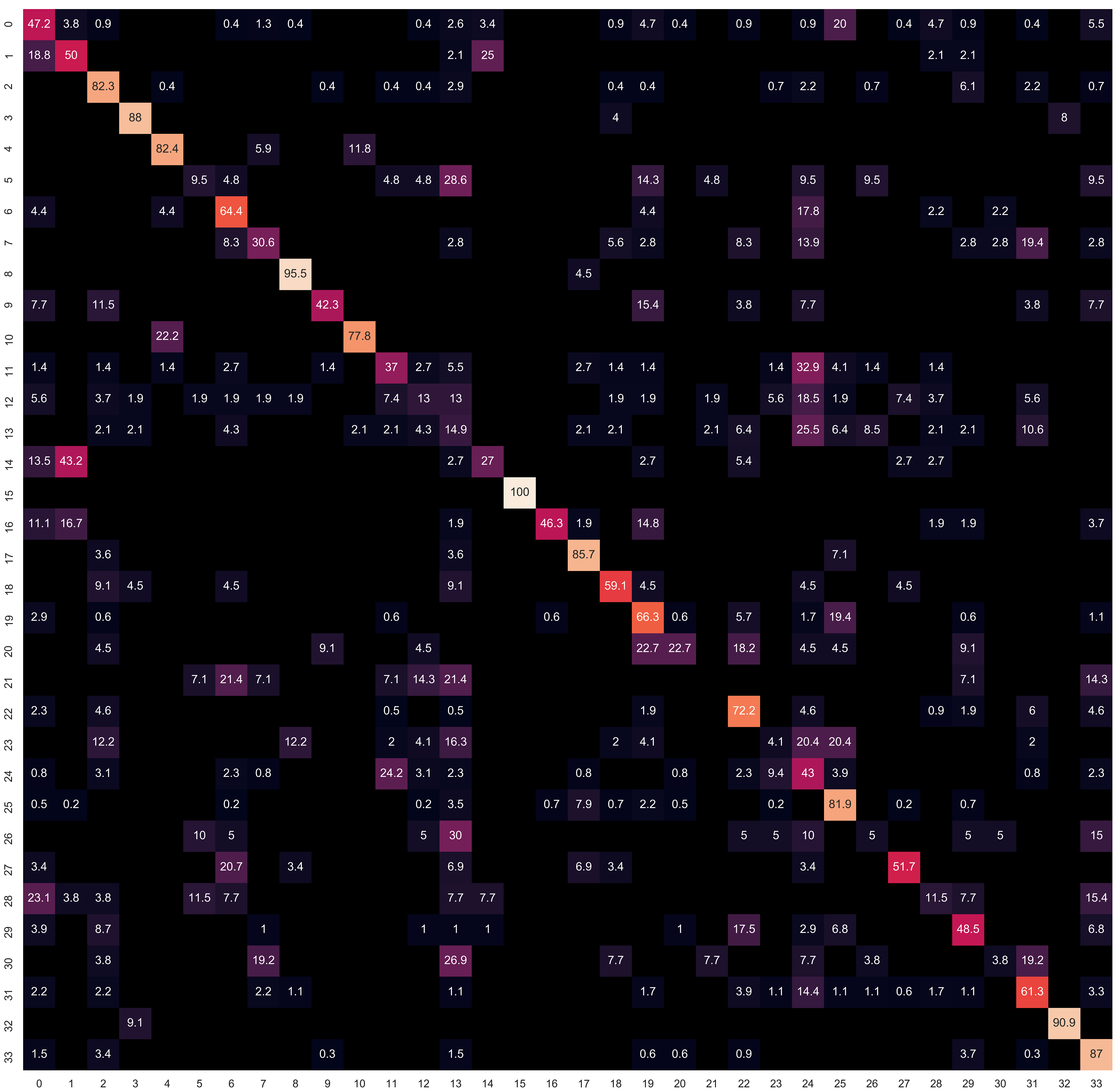}
    \caption{Our CTA-Net's confusion matrix showing 34 fine-grained activities in the Drive\&Act test set.}
\end{figure*}

\begin{figure*}[htbp!]
%\vspace{-5cm}
    \centering
    \subfloat[Reading Magazine (Before, During, After)]{
    \includegraphics[width=0.3\textwidth]{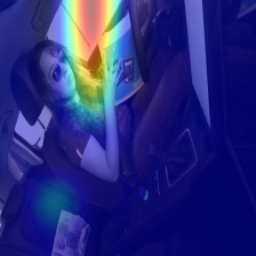}
    \includegraphics[width=0.3\textwidth]{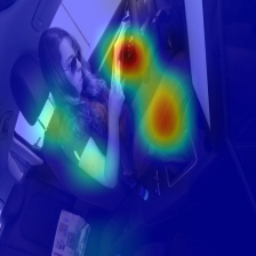}
    \includegraphics[width=0.3\textwidth]{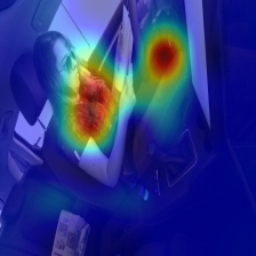}}
    \\
    \subfloat[Reading Newspaper (Before, During, After)]{
    \includegraphics[width=0.3\textwidth]{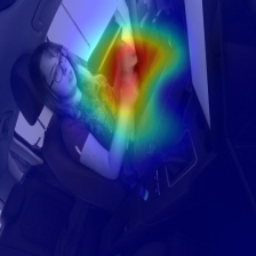}
    \includegraphics[width=0.3\textwidth]{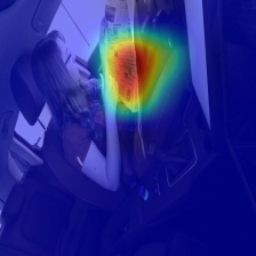}
    \includegraphics[width=0.3\textwidth]{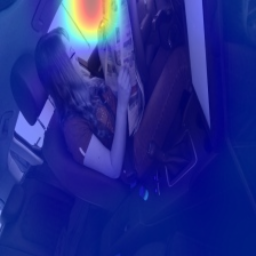}}
\\
    \subfloat[Put on Sunglasses (Before, During, After)]{
    \includegraphics[width=0.3\textwidth]{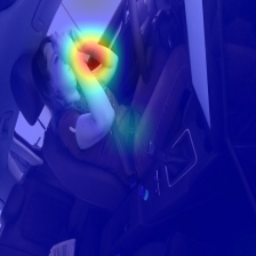}
    \includegraphics[width=0.3\textwidth]{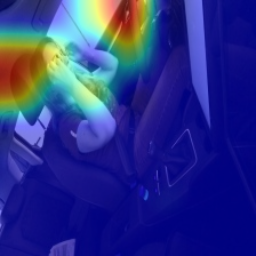}
    \includegraphics[width=0.3\textwidth]{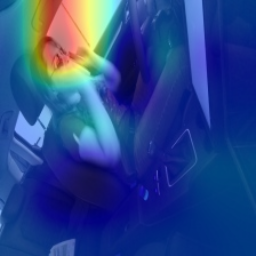}}
    \\
    \subfloat[Put on Seat-belt (Before, During, After)]{
    \includegraphics[width=0.3\textwidth]{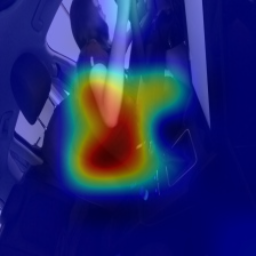}
    \includegraphics[width=0.3\textwidth]{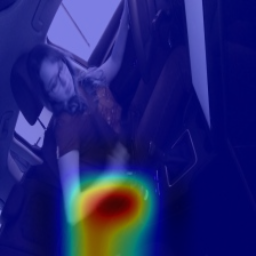}
    \includegraphics[width=0.3\textwidth]{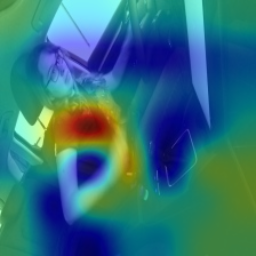}}
    \end{figure*}
    %\clearpage   
    \begin{figure*}[htbp!]\ContinuedFloat
    \subfloat[Working (Before, During, After)]{
    \includegraphics[width=0.3\textwidth]{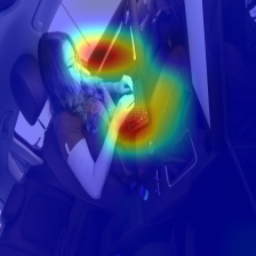}
    \includegraphics[width=0.3\textwidth]{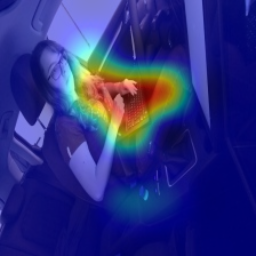}
    \includegraphics[width=0.3\textwidth]{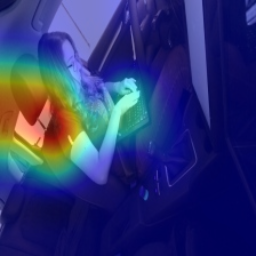}}
    \caption{More examples demonstrating visual explanation of decision from our CTA-Net using class activation map representing our coarse temporal attention of ‘before’ (left), ‘during’ (middle) and ‘after’ (right) segment of 5 different actions }
\end{figure*}

\begin{figure*}[htbp!]
    \centering
    \subfloat[Validation set]{
    \includegraphics[width=0.45\textwidth]{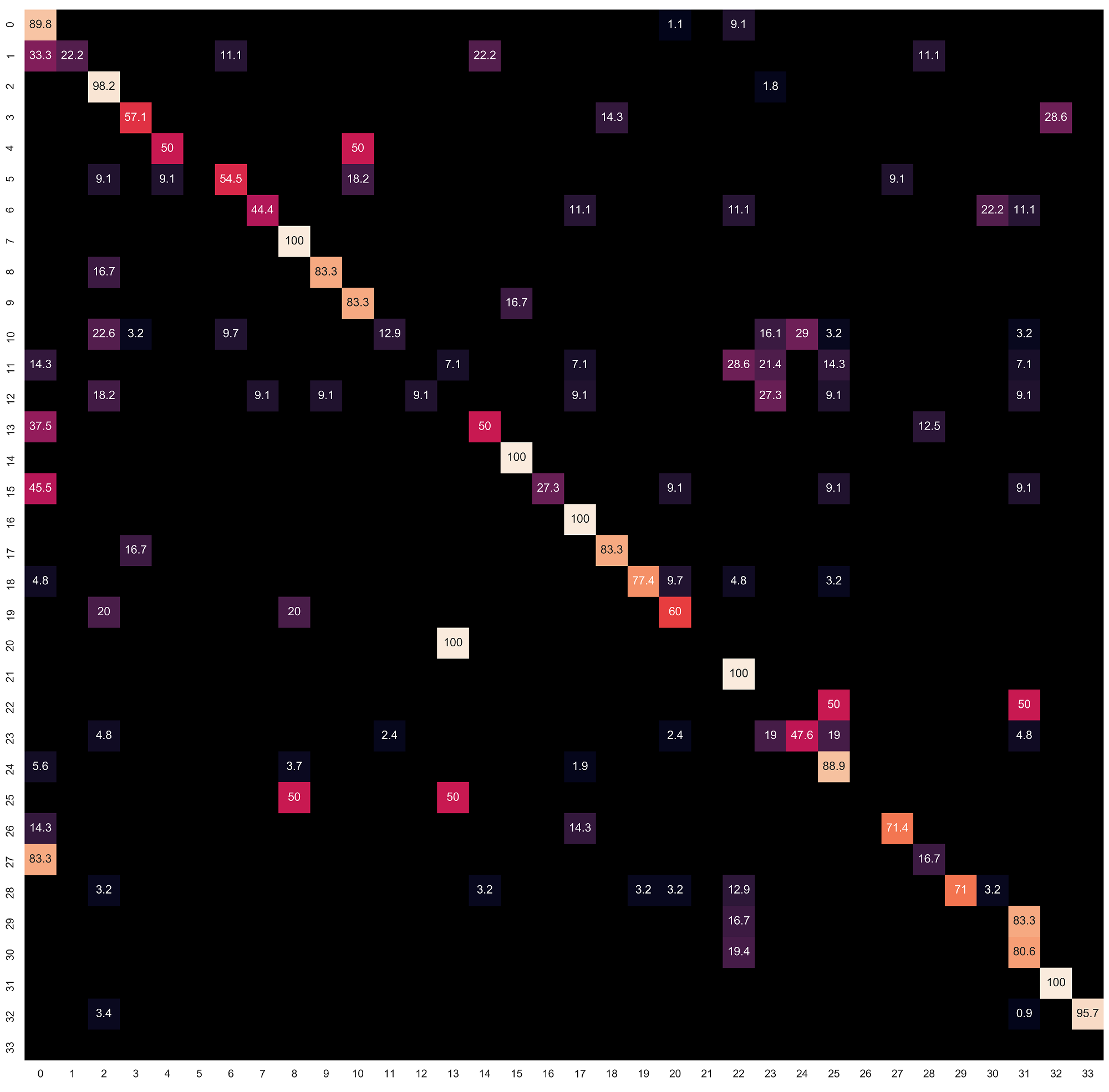}}
    \hfill
    \subfloat[Test set]{
    \includegraphics[width=0.45\textwidth]{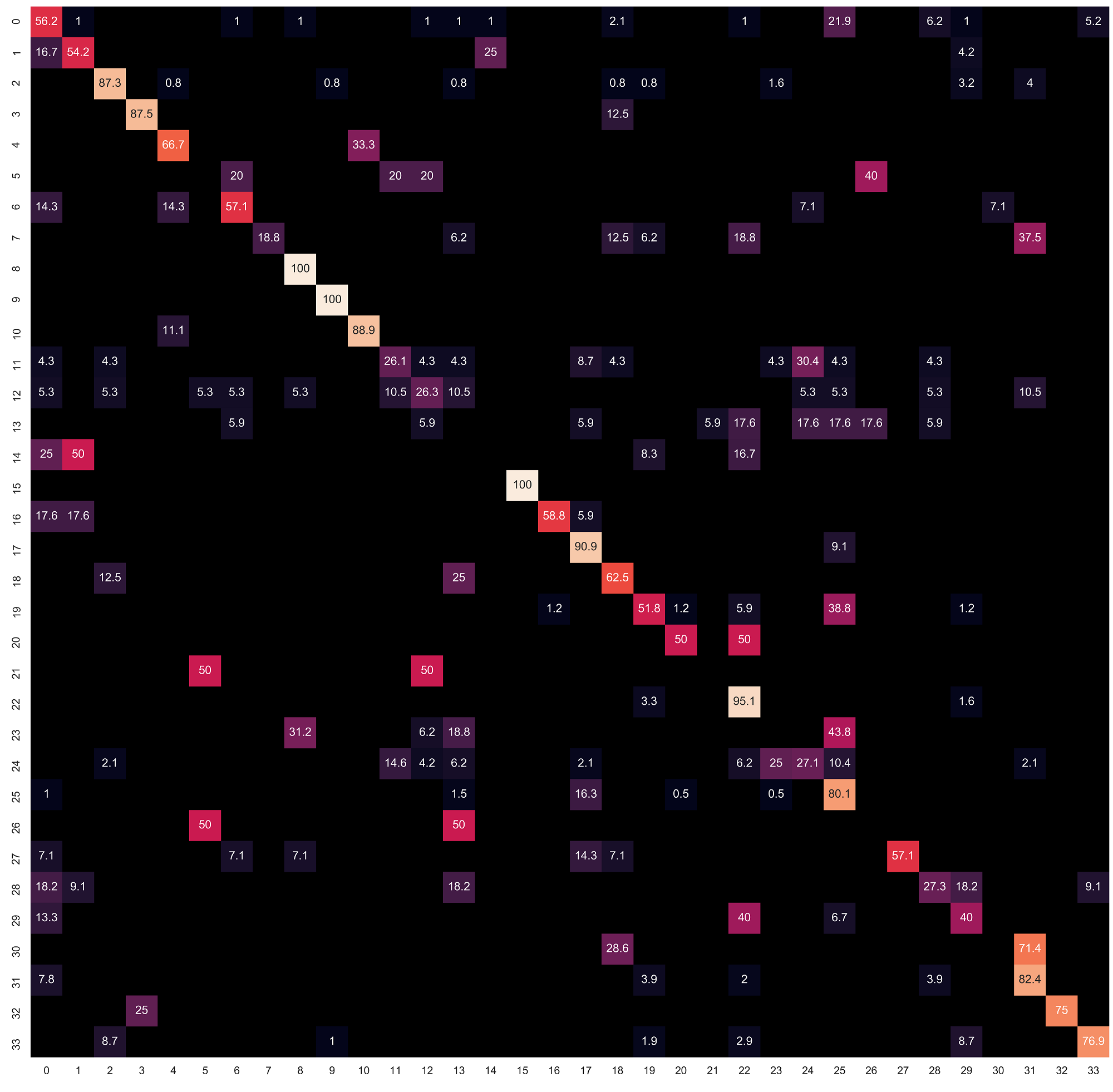}}
    \caption{CTA-Net's confusion matrix of the 34 fine-grained activities using \textbf{split 0} in Drive\&Act dataset}
\end{figure*}
\begin{figure*}[htbp!]
    \centering
    \subfloat[Validation set]{
    \includegraphics[width=0.45\textwidth]{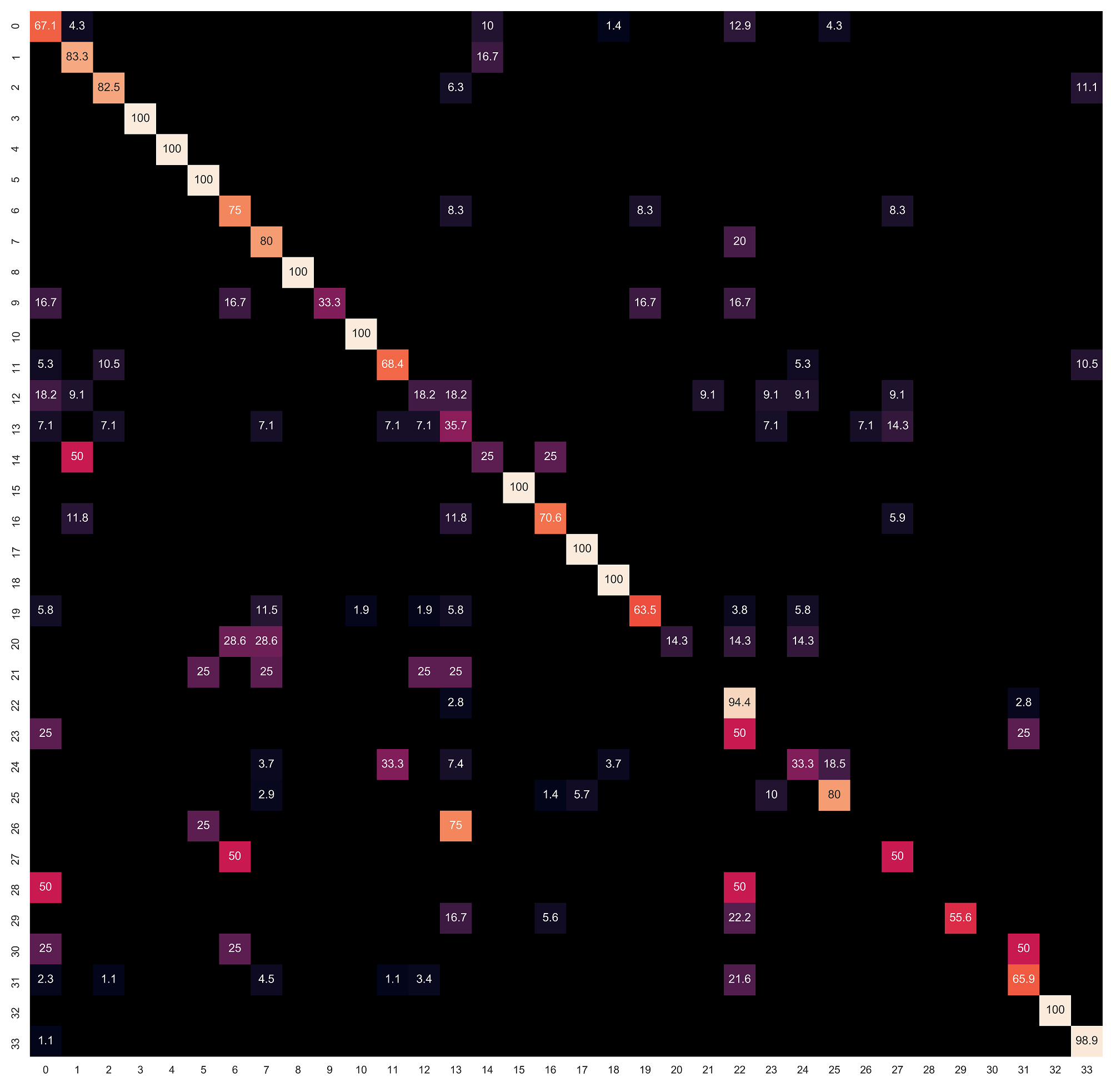}}
    \hfill
    \subfloat[Test set]{
    \includegraphics[width=0.45\textwidth]{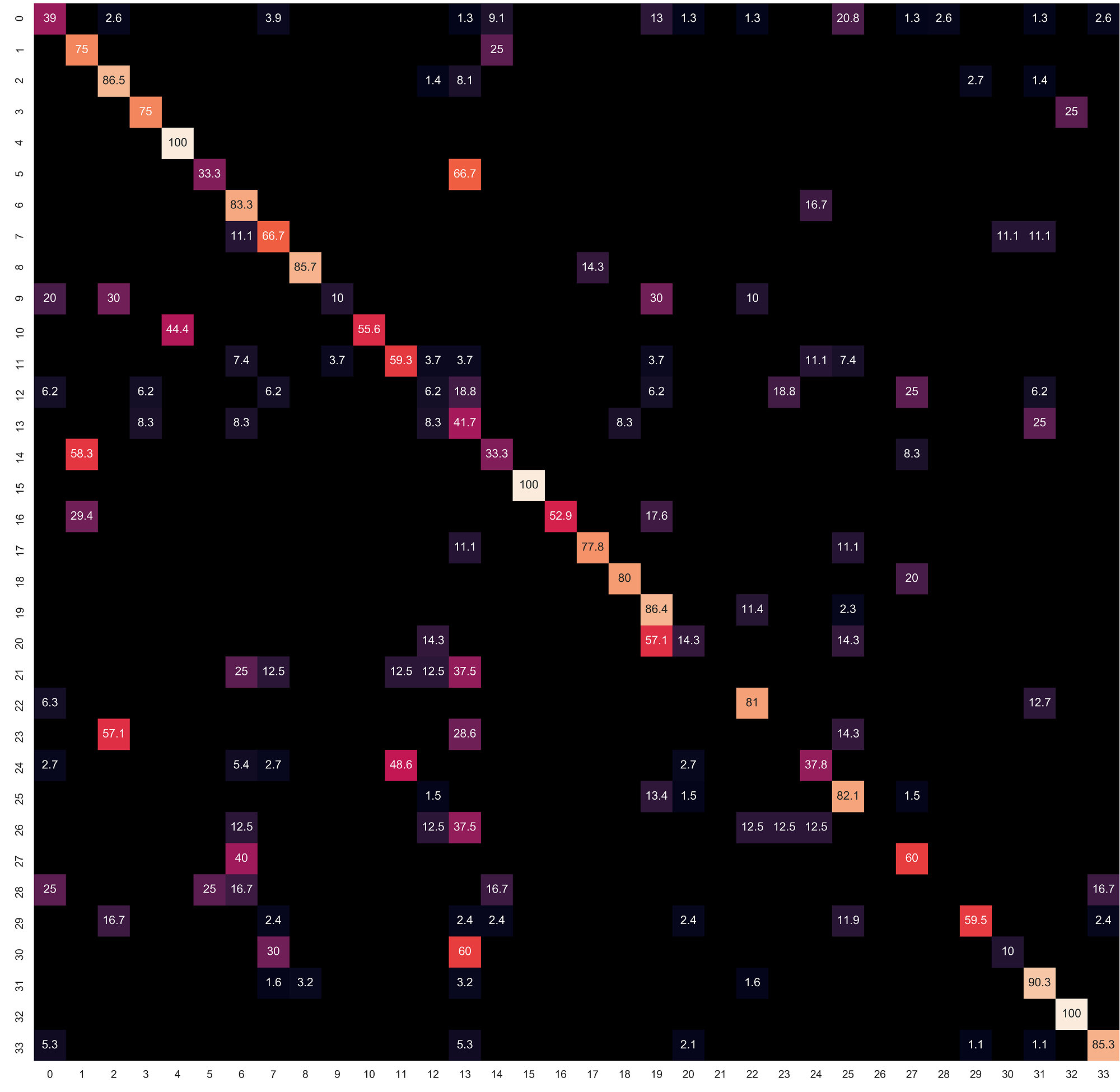}}
    \caption{CTA-Net's confusion matrix of the 34 fine-grained activities using \textbf{split 1} in Drive\&Act dataset}
\end{figure*}
\begin{figure*}[htbp!]
    \centering
    \subfloat[Validation set]{
    \includegraphics[width=0.45\textwidth]{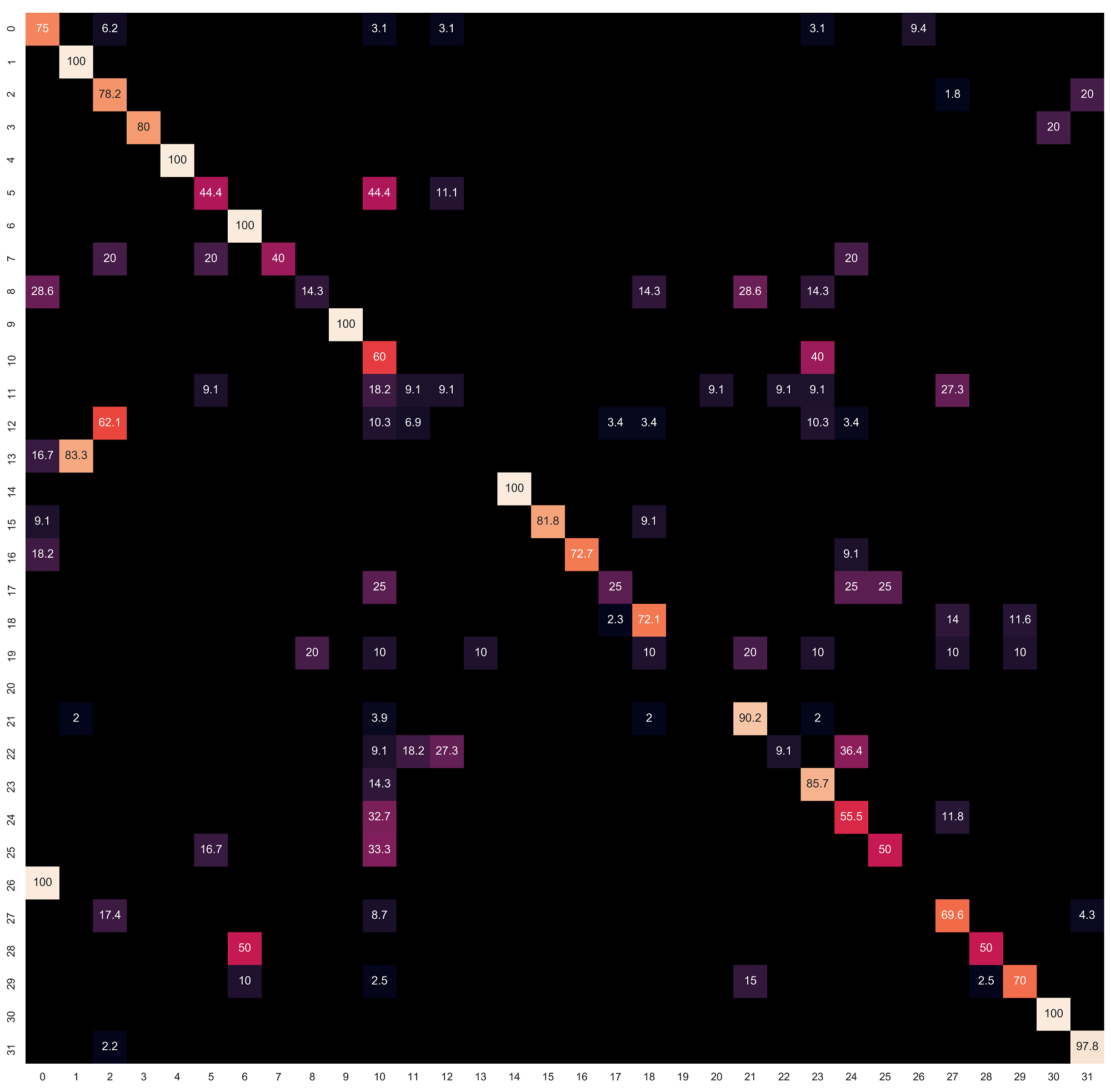}}
    \hfill
    \subfloat[Test set]{
    \includegraphics[width=0.45\textwidth]{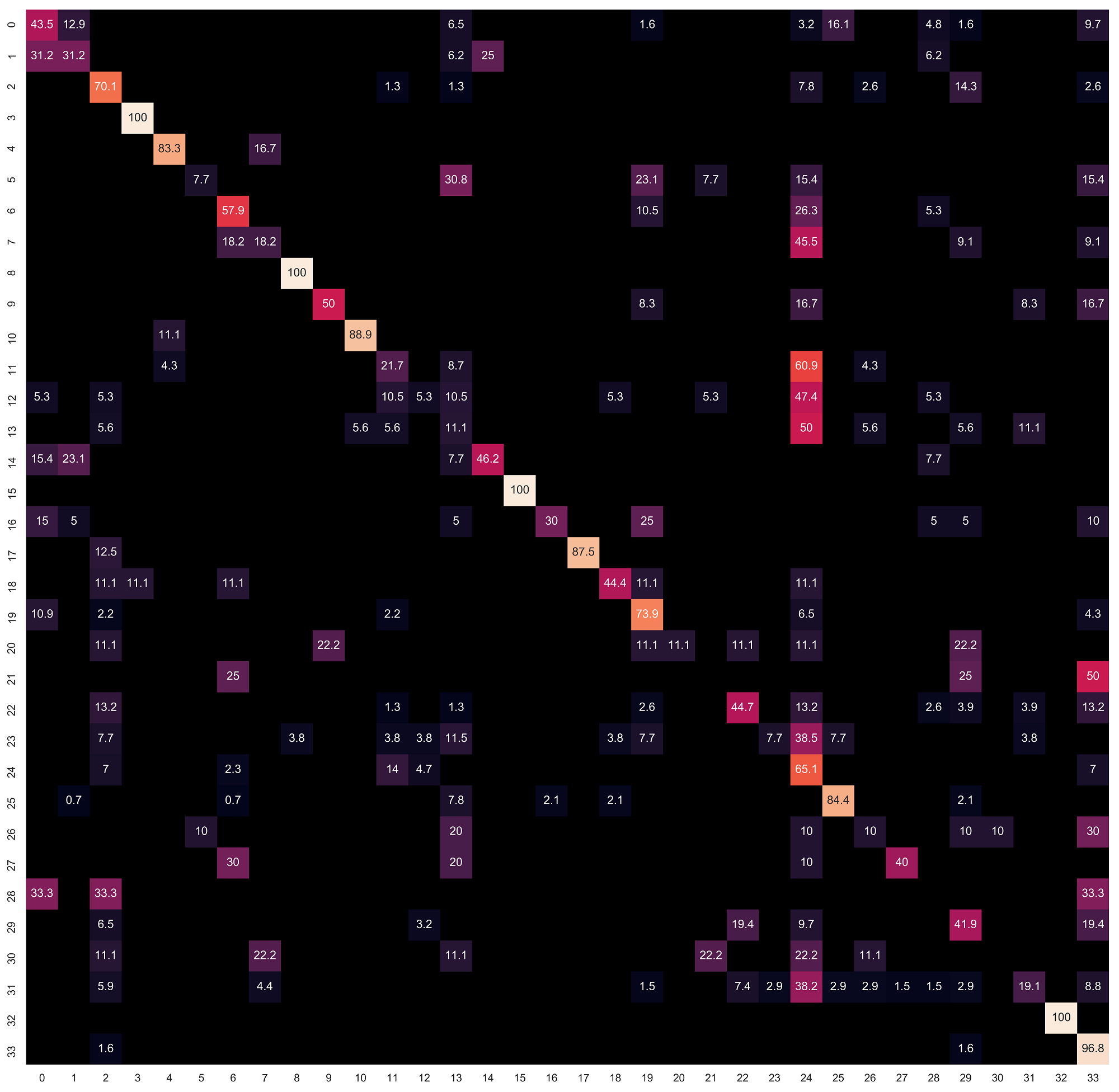}}
    \caption{CTA-Net's confusion matrix of the 34 fine-grained activities using \textbf{split 2} in Drive\&Act dataset}
\end{figure*}
\begin{figure*}[htbp!]
    \centering
    \subfloat[Validation set]{
    \includegraphics[width=0.45\textwidth]{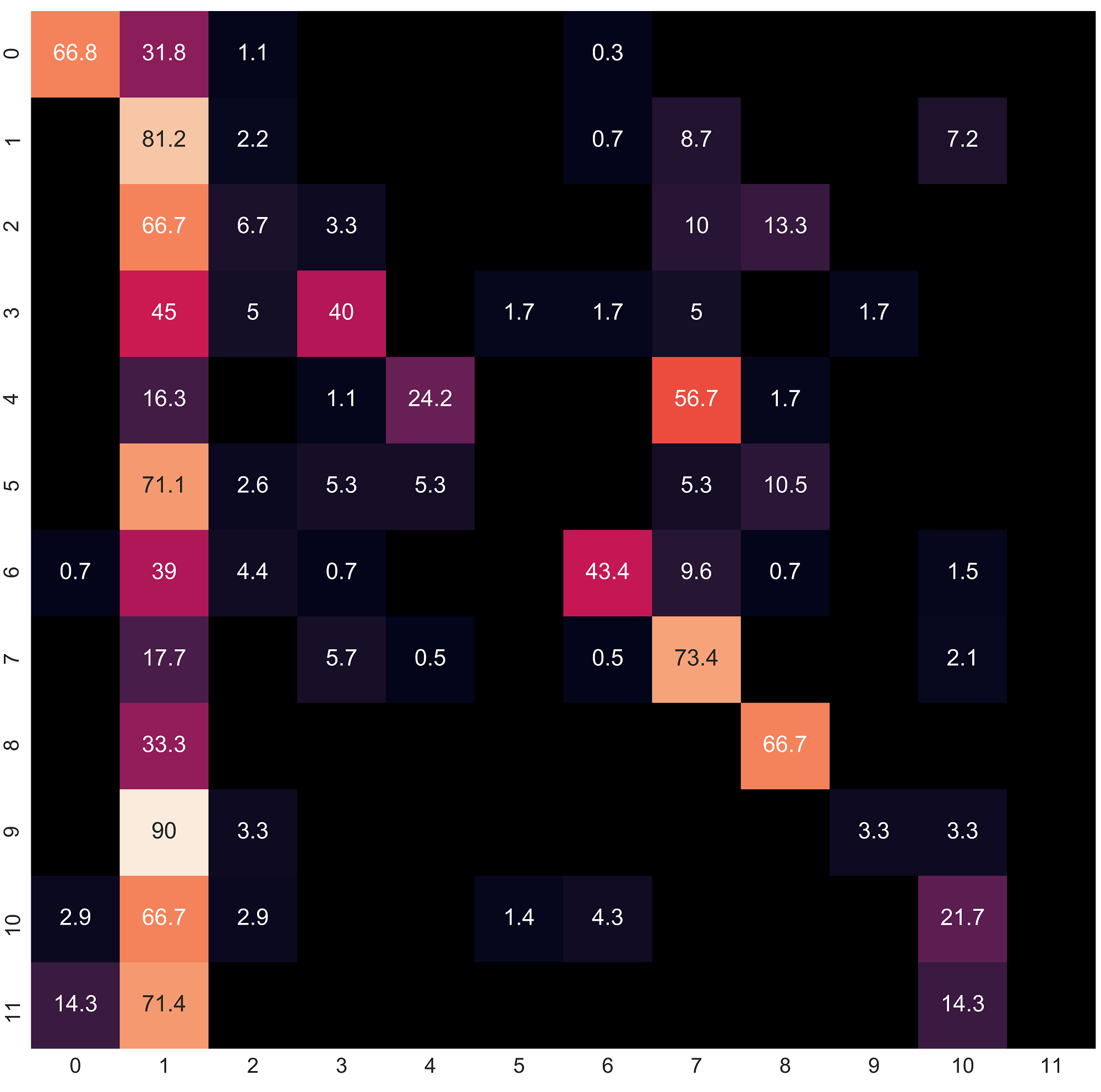}}
    \hfill
    \subfloat[Test set]{
    \includegraphics[width=0.45\textwidth]{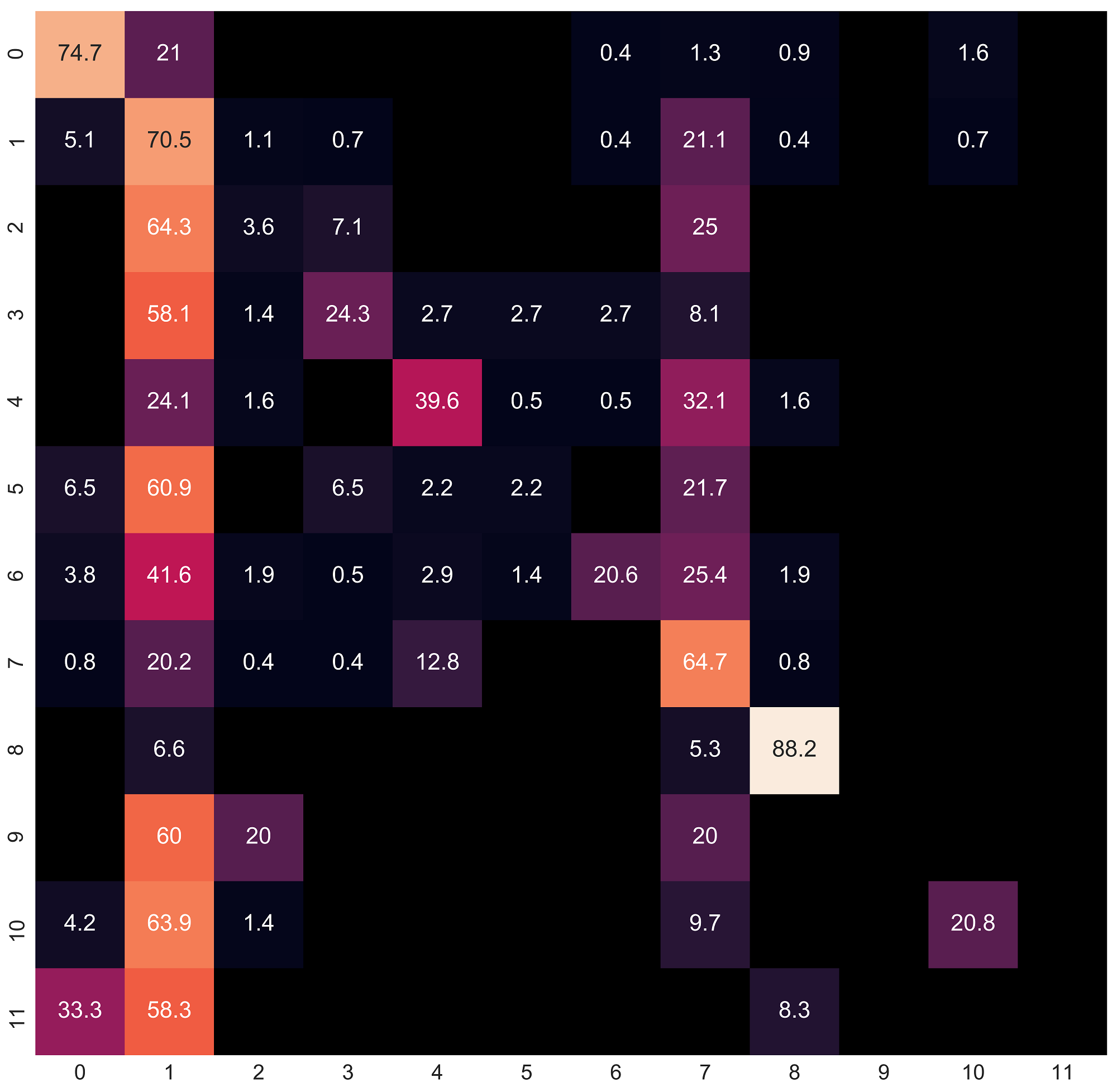}}
    \caption{CTA-Net's confusion matrix of the 12 coarse/scenario tasks using \textbf{split 0} in Drive\&Act dataset}
\end{figure*}
\begin{figure*}[htbp!]
    \centering
    \subfloat[Validation set]{
    \includegraphics[width=0.45\textwidth]{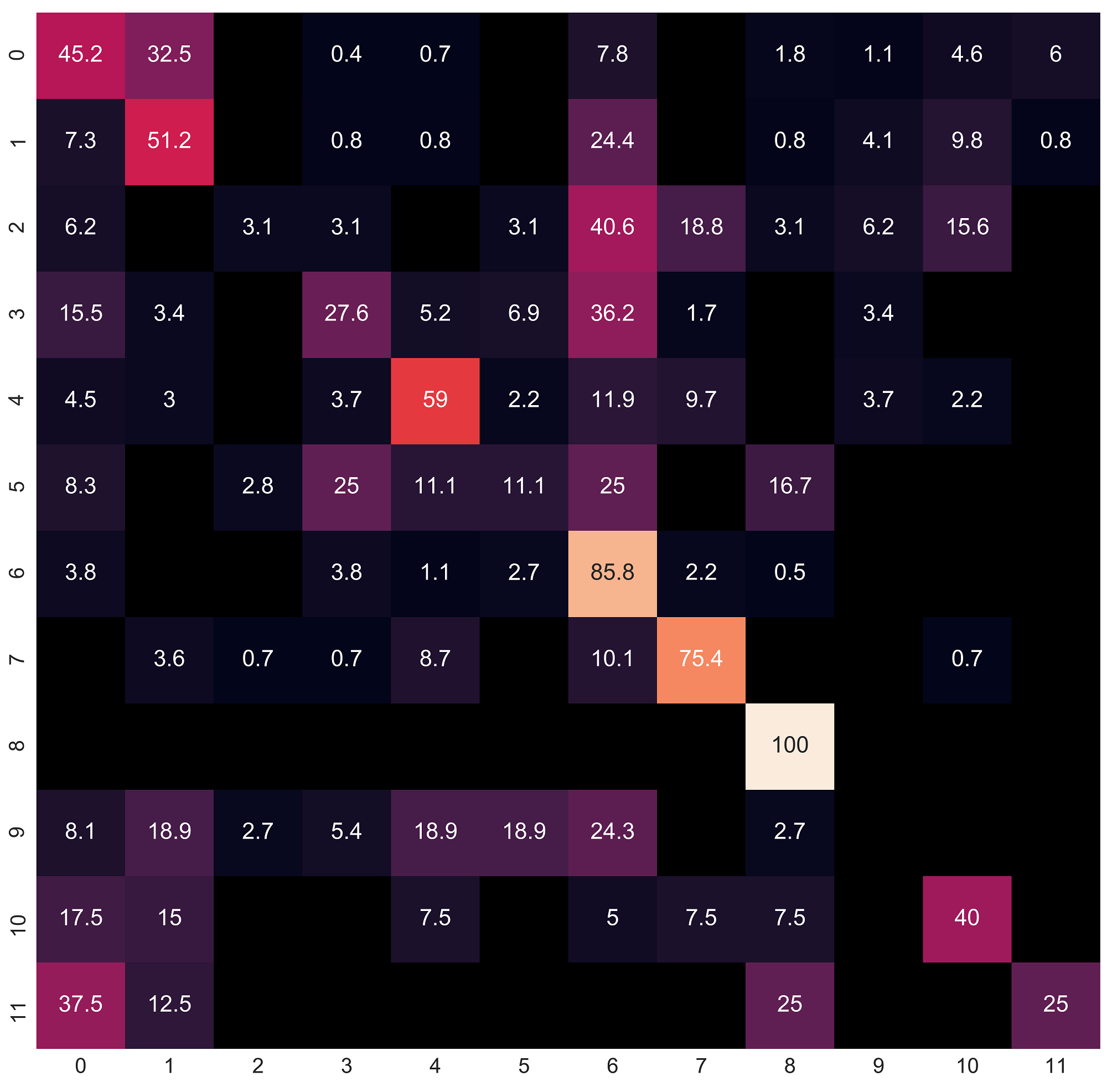}}
    \hfill
    \subfloat[Test set]{
    \includegraphics[width=0.45\textwidth]{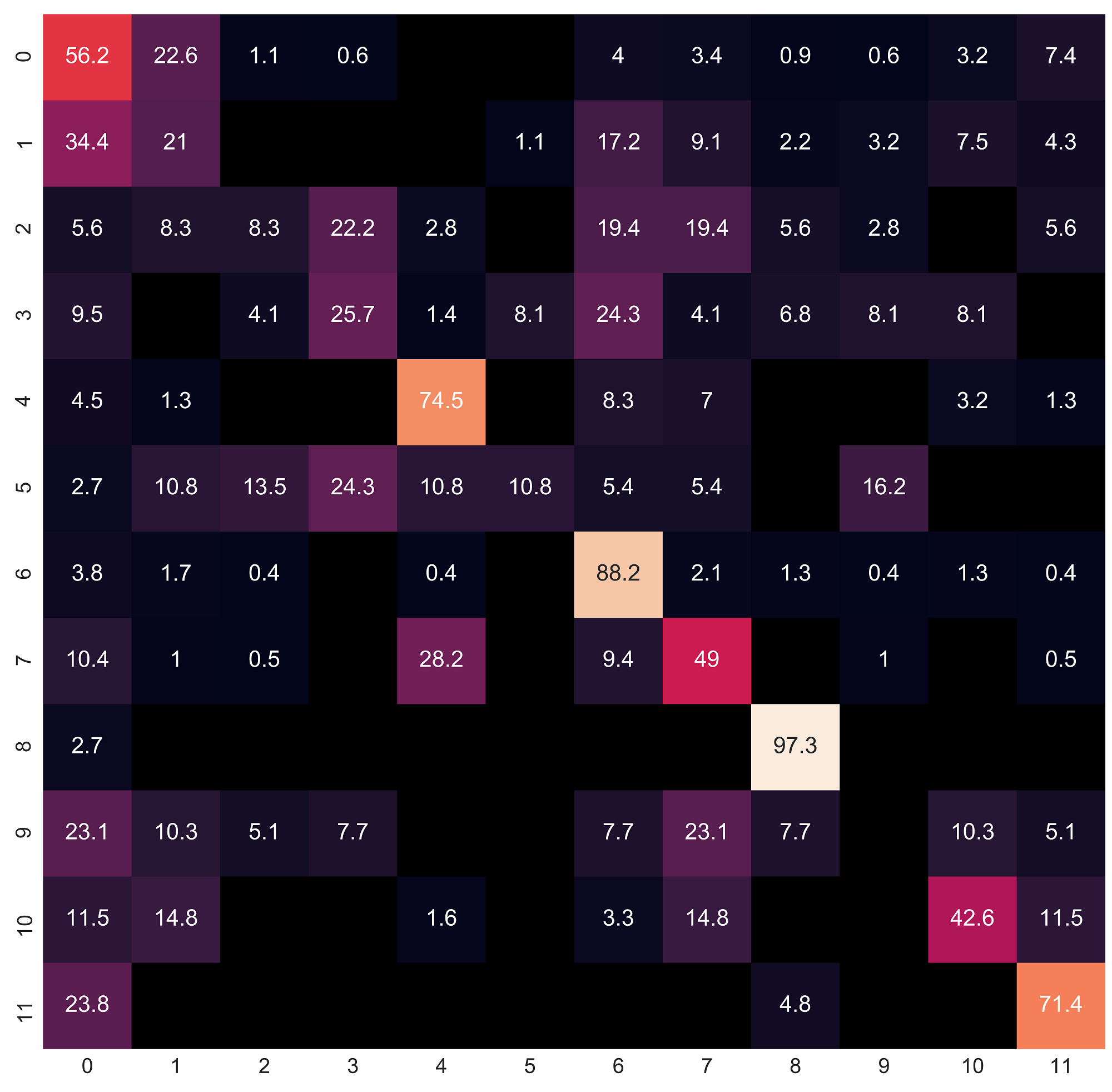}}
    \caption{CTA-Net's confusion matrix of the 12 coarse/scenario tasks using \textbf{split 1} in Drive\&Act dataset}
\end{figure*}
\begin{figure*}[htbp!]
    \centering
    \subfloat[Validation set]{
    \includegraphics[width=0.45\textwidth]{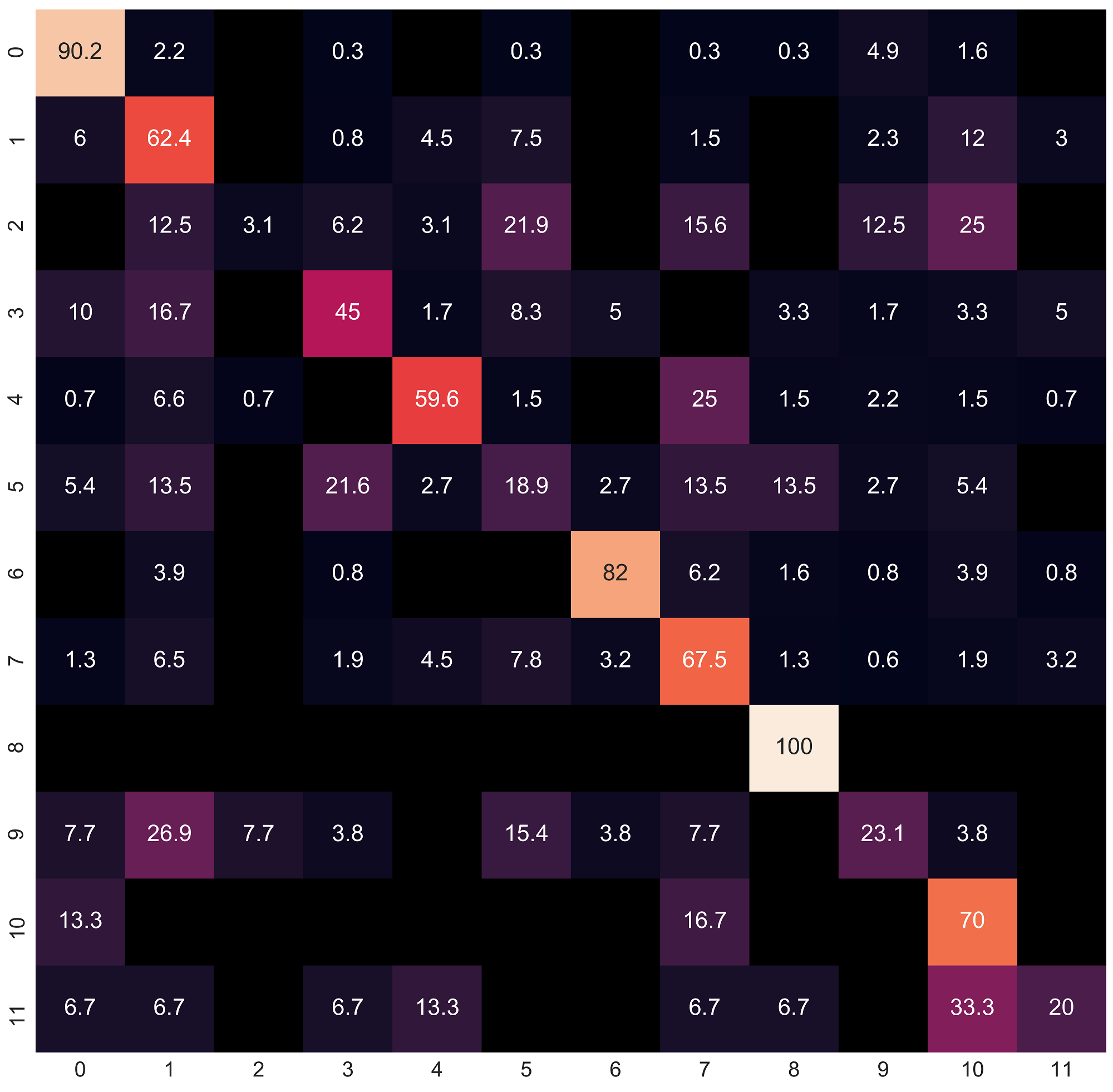}}
    \hfill
    \subfloat[Test set]{
    \includegraphics[width=0.45\textwidth]{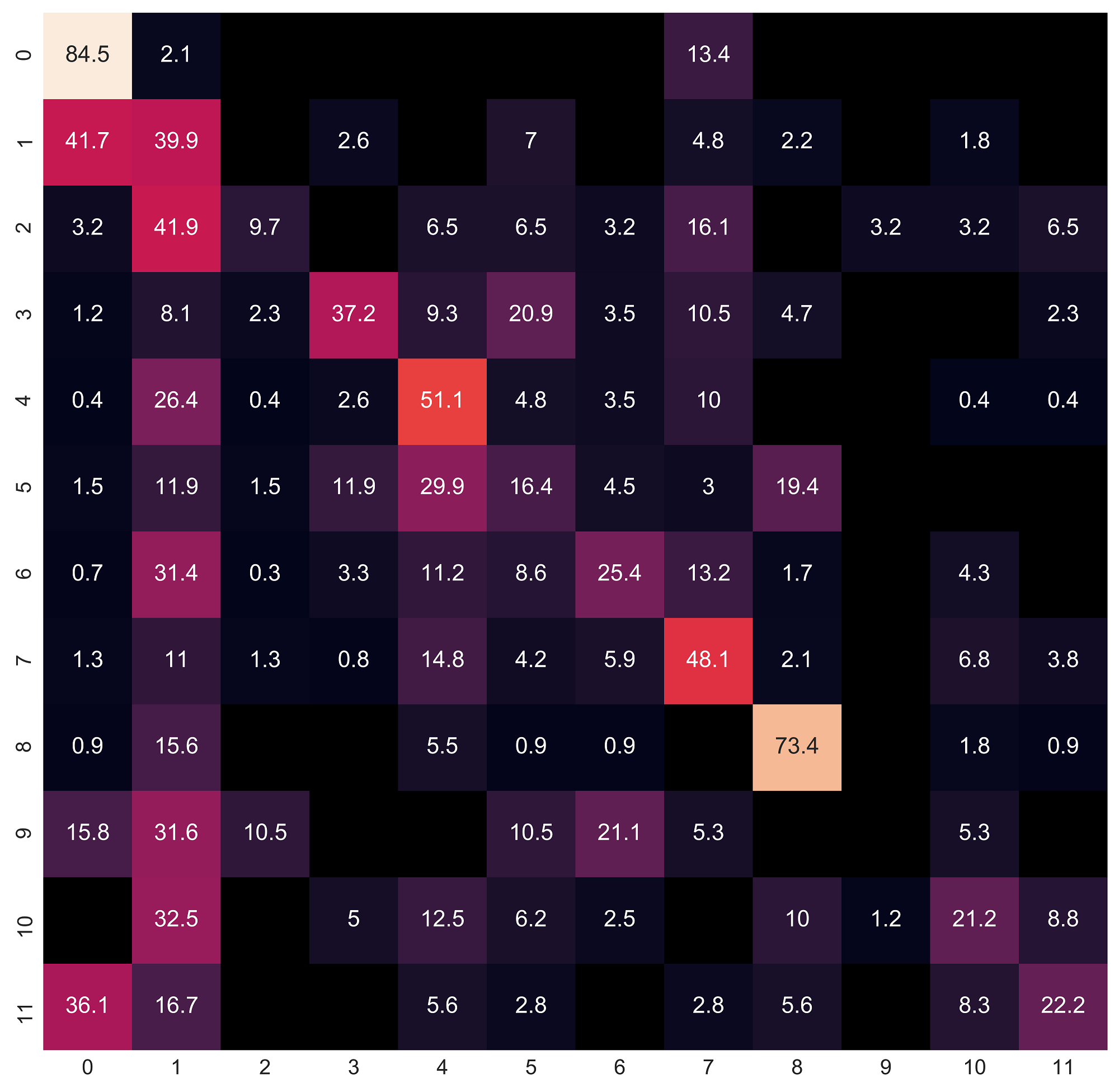}}
    \caption{CTA-Net's confusion matrix of the 12 coarse/scenario tasks using \textbf{split 2} in Drive\&Act dataset}
\end{figure*}

%remove this line to mix references with figures
%\newpage
\clearpage 
%\newline
%continue with rest of the original paper

%===========================================================
{\small
\bibliographystyle{ieee_fullname}
\bibliography{egbib}
}
\end{document}